\documentclass[conference]{IEEEtran}
\IEEEoverridecommandlockouts
\usepackage{cite}
\usepackage{amsmath,amssymb,amsfonts}
\usepackage{algorithmic}
\usepackage{graphicx}
\usepackage{textcomp}
\usepackage[x11names]{xcolor}
\usepackage{tcolorbox}
\usepackage{hyperref}
\usepackage{tabularx}
\usepackage{booktabs}
\def\BibTeX{{\rm B\kern-.05em{\sc i\kern-.025em b}\kern-.08em
    T\kern-.1667em\lower.7ex\hbox{E}\kern-.125emX}}
\begin{document}

\newcommand{\franck}[1]{{\textcolor{red}{ Franck: #1 }}}
\newcommand{\sandeep}[1]{{\textcolor{cyan}{ Sandeep: #1 }}}
\newcommand{\robert}[1]{{\textcolor{blue}{ Robert: #1 }}}
\newcommand{\nes}[1]{{\textcolor{green}{ Nesar: #1 }}}
\newcommand{\neil}[1]{{\textcolor{green}{ Neil: #1 }}}
\newcommand{\tanwi}[1]{{\textcolor{cyan}{ Tanwi: #1 }}}
\newcommand{\murat}[1]{{\textcolor{magenta}{ Murat: #1 }}}
\newcommand{\ian}[1]{{\textcolor{orange}{ Ian: #1 }}}
\newcommand{\eliu}[1]{{\textcolor{red}{ Eliu: #1 }}}

\title{EAIRA: Establishing a Methodology for Evaluating AI Models as Scientific Research Assistants\\ PREPRINT
\thanks{
$^\sigma$Corresponding Authors: cappello@anl.gov, smadireddy@anl.gov
}
}

\author{

\IEEEauthorblockN{Franck Cappello\IEEEauthorrefmark{1}$^\sigma$,
Sandeep Madireddy\IEEEauthorrefmark{1}$^\sigma$,
Robert Underwood\IEEEauthorrefmark{1},
Neil Getty\IEEEauthorrefmark{2},
Nicholas Lee-Ping Chia\IEEEauthorrefmark{2}}
\IEEEauthorblockN{Nesar Ramachandra\IEEEauthorrefmark{3},
Josh Nguyen\IEEEauthorrefmark{6},
Murat Keçeli\IEEEauthorrefmark{3},
Tanwi Mallick\IEEEauthorrefmark{1},
Zilinghan Li\IEEEauthorrefmark{2},}
\IEEEauthorblockN{Marieme Ngom\IEEEauthorrefmark{1},
Chenhui Zhang\IEEEauthorrefmark{10},
Angel Yanguas-Gil\IEEEauthorrefmark{11},
Evan Antoniuk\IEEEauthorrefmark{7},
Bhavya Kailkhura\IEEEauthorrefmark{7},
Minyang Tian\IEEEauthorrefmark{2,5}},
\IEEEauthorblockN{Yufeng Du\IEEEauthorrefmark{2,5},
Yuan-Sen Ting\IEEEauthorrefmark{8},
Azton Wells\IEEEauthorrefmark{3},
Bogdan Nicolae\IEEEauthorrefmark{1},
Avinash Maurya\IEEEauthorrefmark{1},
M. Mustafa Rafique\IEEEauthorrefmark{9},}
\IEEEauthorblockN{Eliu Huerta\IEEEauthorrefmark{2,4,5},
Bo Li\IEEEauthorrefmark{4},
Ian Foster\IEEEauthorrefmark{2,4},
Rick Stevens\IEEEauthorrefmark{2, 4}}

\IEEEauthorblockA{\IEEEauthorrefmark{1} Mathematics and Computer Science Division, Argonne National Laboratory,}
\IEEEauthorblockA{\IEEEauthorrefmark{2} Data Science and Learning Division, Argonne National Laboratory,}
\IEEEauthorblockA{\IEEEauthorrefmark{3} Computational Science Division, Argonne National Laboratory,}
\IEEEauthorblockA{\IEEEauthorrefmark{11} Applied Materials Division, Argonne National Laboratory,}
\IEEEauthorblockA{\IEEEauthorrefmark{4} Department of Computer Science, The University of Chicago,}
\IEEEauthorblockA{\IEEEauthorrefmark{6} University of Pennsylvania,
\IEEEauthorrefmark{7} Lawrence Livermore National Laboratory,}
\IEEEauthorblockA{\IEEEauthorrefmark{8} The Ohio State University,
\IEEEauthorrefmark{9} Rochester Institute of Technology,
\IEEEauthorrefmark{10} Massachusetts Institute of Technology.}
}

\maketitle

\begin{abstract}

Recent advancements have positioned AI, and particularly Large Language Models (LLMs) as transformative tools for scientific research, capable of addressing complex tasks that require reasoning, problem-solving, and decision-making. Their exceptional capabilities suggest their potential as scientific research assistants, but also highlight the need for holistic, rigorous, and domain-specific evaluation to assess effectiveness in real-world scientific applications. This paper describes a multifaceted methodology for Evaluating AI models as scientific Research Assistants (EAIRA) 
developed at Argonne National Laboratory. 
This methodology incorporates four primary classes of evaluations.
1) Multiple Choice Questions to assess factual recall; 2) Open Response 
to evaluate advanced reasoning and problem-solving skills; 3) Lab-Style Experiments involving detailed analysis of capabilities as research assistants in controlled environments; and 4) Field-Style Experiments to capture researcher-LLM interactions at scale in a wide range of scientific domains and applications. These complementary methods enable a comprehensive analysis of LLM strengths and weaknesses with respect to their scientific knowledge, reasoning abilities, and adaptability. Recognizing the rapid pace of LLM advancements, we designed the methodology to evolve and adapt so as to ensure its continued relevance and applicability. This paper describes the methodology's state at the end of February 2025. Although developed within a subset of scientific domains, the methodology is designed to be generalizable to a wide range of scientific domains. 
\end{abstract}


\section{Introduction}

\begin{table*}[htb] 
    \centering
        \caption{Our proposed methodology for evaluating LLMs as scientific assistants combines four complementary techniques, listed in columns 2--5 below, to assess their capabilities. \textcolor{purple}{Purple text} indicates \textbf{prior} contributions by the authors, \textcolor{blue}{blue text} \textbf{new} contributions in this paper, and  black text methods adapted from existing work that we include for a complete approach.
        }
    \label{fig:multi_facet}
    \hrule
    \includegraphics[width=\textwidth, clip]{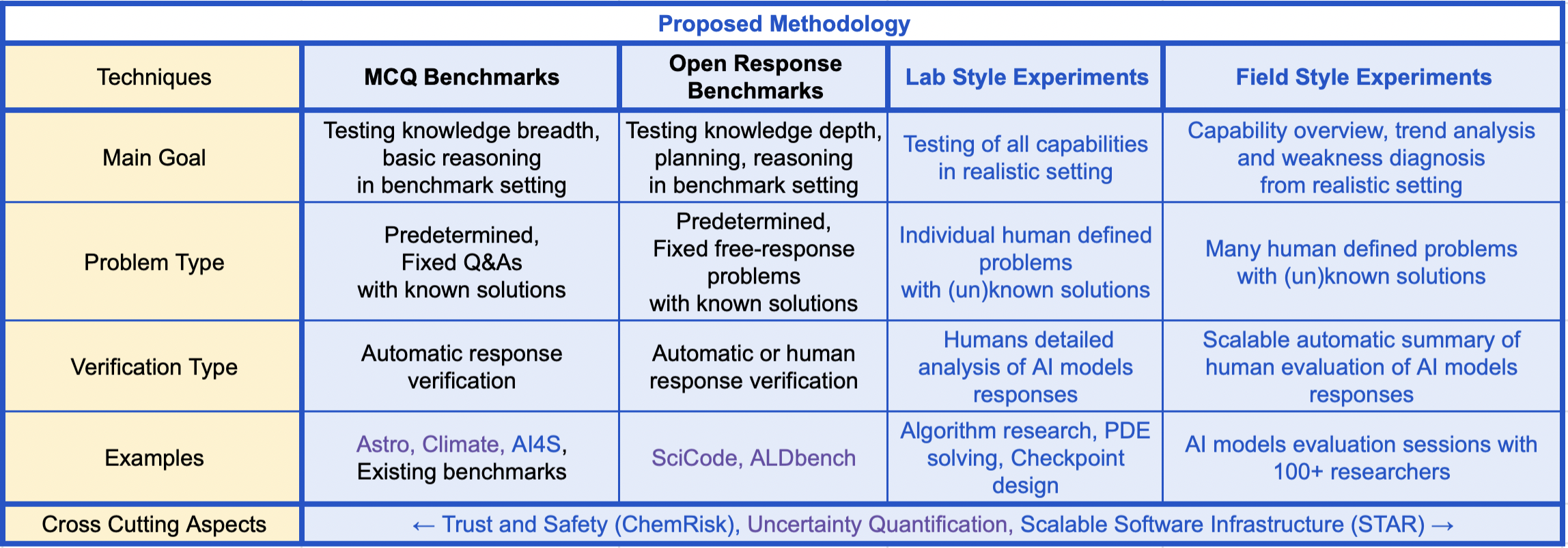} 
    \vspace{-8mm}
\end{table*}

Recent advances in Large Language Models (LLMs) have greatly broadened conceptions of what AI may be able to accomplish in the near future. Models such as OpenAI's GPT O1 \cite{openaiGPT4TechnicalReport2024}, Google's Gemini \cite{geminiteam2024gemini}, and Anthropic's Claude \cite{anthropic2024claude3} 
are transforming traditional natural language understanding (NLU) tasks like summarization, information extraction, translation, and classification with enhanced contextual depth and adaptability. They are also exhibiting promising potential \textit{beyond} NLU, with measurable progress on tasks such as mathematical problem solving, multi-step reasoning, and symbolic logic---and achieving significant milestones such as passing the Uniform BAR exam and medical licensing exams~\cite{varanasiGPT4CanAce2023}. Such achievements highlight their potential to emulate abstraction, logical deduction, and domain-specific expertise. This evolution from NLU towards addressing complex, domain-specific challenges with minimal human guidance has propelled LLMs into a pivotal role for next-generation AI systems, positioning them as a cornerstone technology in the quest toward more general-purpose AI, and potentially, artificial general intelligence (AGI).

Building on these advancements, scientists are now beginning to assess \textit{separately} the suitability and potential impact of LLMs on a wide variety of specific tasks within specific fields of research and discovery~\cite {ai4science2023impactlargelanguagemodels}. This work has led to exciting demonstrations of LLMs as transformative tools for such tasks predicting molecular properties \cite{liu2024moleculargpt}, uncovering genomic patterns \cite{benegas2025genomic}, interpreting astrophysical data \cite{ting2024astromlab}, solving mathematical problems \cite{hendrycksmath2021}, and even creating and manipulating tools for simulations and analysis \cite{schick2023toolformer}. 

These developments have also led scientists to envision the use of LLMs and transformers as research assistants that can not only
automate individual research tasks but also engage with scientific problems in depth by taking advantage of growing multi-step reasoning skills that complement their expanding contextual understanding. This vision suggests a new \emph{holistic} approach in which LLMs interface with relevant tools, operate (quasi-)autonomously on research challenges, identify relevant literature, summarize findings, propose experimental designs, and even autonomously run, and generate insights from, physical and computational experiments \cite{liu2024towards, ma2024llm}.

We identify \textbf{two main challenges} that must be addressed before LLMs can be broadly adopted by the scientific community as effective and trustworthy research assistants. 
\textbf{First}, researchers need ways to measure and evaluate LLM capabilities in the different stages and tasks of the scientific research process. Such evaluations can both guide LLM applications and integration with other tools, and provide benchmarks for developers to improve their LLMs and supporting systems. \textbf{Second}, as with other research tools and techniques, researchers need ways to assess confidence in the results produced, in order to decide whether or not they are trustworthy. A comprehensive, rigorous, accurate, transparent and community-approved evaluation methodology is necessary to address these two challenges.

This paper introduces work undertaken at Argonne National Laboratory within the AuroraGPT project to develop a research methodology that addresses these two challenges. The paper makes \textbf{three main contributions}: (i) a holistic methodology for LLM evaluation; (ii) two novel evaluation techniques (lab-style and field-style experiments); and (iii) improvements in existing state-of-the-art evaluation techniques for the specific role of research assistant.

\textbf{First}, we propose an overarching methodology for assessing the scientific knowledge, skills, and safety of AI models. As shown in \autoref{fig:multi_facet}, this methodology encompasses four complementary techniques: 1) Multiple Choice Question (MCQ) Benchmarks, which measure factual recall and reasoning capabilities in structured formats to provide fast assessment of a model's breadth of knowledge; 2) Open Response Benchmarks, which test a model’s ability to generate detailed open-ended responses and write or debug code for computational tasks giving a fast but more in-depth analysis of knowledge; 3) Lab-style Experiments, simulating various tasks in the end-to-end research process to assess model performance on those tasks and thus to provide understanding of real-world strengths and weaknesses; and 4) Field-style Experiments, capturing real-world interactions at scale to analyze user needs, model strengths, and broad capability trends, and to diagnose areas and sources of weakness in realistic scenarios. We also consider trustworthiness, i.e., alignment with ethical and safety standards, and discuss the Software Infrastructure required to implement these methodologies effectively.

\textbf{Second}, our application of ``lab-style'' and ``field-style'' techniques to LLM evaluation represents a novel approach when conducted at this scale and with such a diversity of topics with practicing scientists. These techniques go beyond existing testing methodologies: they assess in real situations the suitability of LLMs for open and unstructured problems which are both common in research and difficult to assess with either MCQs or open-response questions.

\textbf{Third}, we present improvements to techniques used within the community, including a multi-domain AI for science benchmark, called ``AI4S,'' and the Skills, Trust, and Reliability (STaR) evaluation framework, a scalable software evaluation infrastructure. 
We also summarize and contextualize the research done by our team in domain-specific benchmarks, open response benchmarks, and uncertainty quantification, and note where we employ tools from other teams to complete our holistic evaluation methodology.

Together, these efforts aim to establish a robust methodology for evaluating 
the capabilities of LLMs as trusted scientific assistants.
In the following sections, the collection of evaluations and scoring was based on voluntary participation of researchers and contributors.  

The following sections describe the related work and the details of our proposed methodology. The last section discusses next steps.

\section{Related Work}
Research on LLM evaluation encompasses various techniques relevant for the evaluation of LLMs as research assistants. Here we discuss that work and note gaps that our proposed methodology attempts to fill.

\subsection{Multiple-choice Question (MCQ) Benchmarks}
MCQ benchmarks offer a structured framework for assessing LLM performance across various domains. Notable examples are Massive Multitask Language Understanding (MMLU) \cite{hendryckstest2021} and MMLU-PRO \cite{mmlupro}, which evaluate general knowledge and reasoning in more than 50 subjects, including humanities, sciences and engineering. In the realm of mathematical reasoning, GSM8K \cite{cobbe2021training}, GSM1K \cite{zhang2024careful}, and MATH \cite{hendrycksmath2021} are prominent. 
GSM8K and GSM1K addresses grade-school level problems, while MATH focuses on high-school and competition-level questions. Both benchmarks have been enhanced with multiple-choice adaptations to streamline evaluation and minimize ambiguity in model outputs \cite{zhang2024multiple}. Other significant MCQ benchmarks include ARC \cite{clark2018think}, which tests scientific reasoning, and HellaSwag \cite{zellers2019hellaswag}, which challenges models with complex commonsense reasoning scenarios that, while easy for humans, are especially hard for state-of-the-art models.
In the field of chemistry, MCQ benchmarks include MoleculeQA \cite{lu_moleculeqa_2024}, which comprises 61,574 MCQs, each with three distractors, focusing on factual information about molecules; MolPuzzle \cite{guo_can_2024}, a multimodal benchmark with over 23,000 question-answer pairs, structured with interlinked sequential sub-tasks, each providing multiple choices; and ChemBench \cite{mirza2024chembench}, with over 2700 questions, primarily MCQs.
Few of the many other MCQ benchmarking efforts in the literature are science-domain focused and validated, and as LLM capabilities improve, there is an increasing need to generate more difficult questions and leverage synergies between domain expertize and LLM judges. The multi-domain benchmark AI4S that we present below is an attempt towards that end.

\subsection{Open Response Benchmarks}
While MCQ benchmarks restrict responses to predefined options, 
Open Response benchmarks require that LLMs produce detailed, unconstrained outputs that can be evaluated for coherence, accuracy, and relevance. Notable examples include NarrativeQA \cite{kovcisky2018narrativeqa}, which challenges models to generate summaries or interpret longer narratives, and HotpotQA \cite{yang2018hotpotqa}, which demands multi-hop reasoning with synthesized answers derived from multiple sources. Similarly, HybridQA \cite{chen2020hybridqa} combines textual and tabular data, requiring LLMs to provide coherent and comprehensive answers. For mathematical reasoning, GSM8K \cite{cobbe2021training} and MATH \cite{hendrycksmath2021} test model ability to solve complex, multi-step problems with free-form solutions. 
In chemistry, benchmarks have been developed to assess LLM open-response capabilities. ChemistryQA \cite{wei_chemistryqa_2020} comprises 4500 complex questions that require reasoning and calculations, evaluating model ability to generate detailed, accurate responses. ChemLLMBenchmark \cite{guo2023can} consists of eight practical chemistry tasks that necessitate understanding, reasoning, and explanatory skills, with evaluations focusing on the quality and depth of model-generated answers. The open-domain TOMG-Bench \cite{li_tomg-bench_2024} molecule generation benchmark comprises tasks such as molecule editing, optimization, and customized generation, each requiring models to produce specific molecular structures or modifications based on textual descriptions.

Recent advancements have expanded the scope of open-response benchmarks to address specific evaluation challenges and domains. The Open-LLM-Leaderboard (OSQ-bench) \cite{open-llm-leaderboard} transitions from multiple-choice formats to open-style questions, eliminating issues like selection bias and random guessing, while emphasizing models’ ability to generate coherent, contextually accurate answers. FrontierMath \cite{frontiermath} presents hundreds of exceptionally challenging mathematics problems that require models to generate detailed solutions, testing advanced reasoning and problem-solving skills. Similarly, Humanity's Last Exam \cite{humanityslastexam} crowdsources complex questions from experts across fields to evaluate how closely LLMs approximate expert-level capabilities, highlighting their potential and limitations in addressing real-world challenges.

A difficulty with Open Response benchmarks is that evaluating their responses is inherently challenging due to their unstructured nature, requiring time-intensive analysis, subjective interpretation, and careful management of biases and data overload. These challenges are closely tied to uncertainty quantification (UQ), as the variability in interpretations and outcomes necessitates robust techniques to quantify and mitigate uncertainty in the evaluation process, ensuring reliable and consistent insights. In addressing these challenges, it is also important to keep track of multiple model versions and to use them consistently \cite{EvoStore-HPDC24}, so as to enable reproducibility \cite{RECUP-REWORDS23}.

\subsection{Lab-Style Experiments}

Laboratory-style Experiments with LLMs involve controlled settings in which researchers systematically evaluate model performance on specific tasks, enabling precise measurement of capabilities and limitations. In chemistry, for instance, \cite{boiko2023autonomous} developed Coscientist, which employs LLMs to plan and execute experimental procedures based on simple human prompts, and evaluated its performance by assigning it the task of identifying synthetic procedures for seven molecules of varying complexity. Similarly, \cite{chen2024autotamp} proposed an approach that combines LLMs with task and motion planning to translate natural language instructions into robot-executable plans, evaluating their system through simulations in a box-packing domain. In behavioral strategy research, \cite{albert2024reproducing} reproduced human laboratory experiments using LLMs and compared their performance to human participants to analyze the extent to which LLMs can emulate human decision-making processes. These laboratory-style experiments provide valuable insights into applications and can inform the development of more advanced AI systems. However, comprehensive end-to-end evaluations of LLMs on scientific tasks, similar to human performance assessments, remain scarce. Such evaluations are crucial to understanding how LLMs can mimic or augment human researchers in tackling complex scientific challenges. 
\subsection{Field-Style Experiments}
Field-style Experiments, also referred to as ``in-the-wild" studies, involve observing and analyzing user interactions with LLM in real-world settings. This approach contrasts with controlled Lab-style Experiments by capturing non-predefined user interactions with LLMs, providing valuable insights into how LLMs perform across diverse, unstructured tasks. Recent studies have leveraged this methodology to assess various aspects of LLM performance. For instance, WildBench \cite{lin2024wildbench} is designed to evaluate LLMs using real-world user interactions, enabling a comprehensive analysis of model capabilities in practical scenarios. Similarly, HaluEval-Wild \cite{zhu2024haluevalwildevaluatinghallucinationslanguage} was designed to evaluate hallucinations in LLMs by collecting challenging user queries from real-world interactions. Shen et al. \cite{shen2024donowcharacterizingevaluating} conducted a study characterizing and evaluating user-LLM interactions, providing insights into user behavior and model performance and highlight the importance of understanding user needs and expectations to improve LLM utility and user satisfaction. While such analyses are currently lacking in scientific domains, developing field-style experiments specifically for scientific research presents a significant opportunity to provide critical insights into how researchers interact with AI models, thereby enhancing the creation of AI assistants tailored to scientific inquiry.

\subsection{Safety Evaluation}
In addition to the general-purpose evaluations discussed above, comprehensive evaluations must rigorously assess alignment with ethical standards, robustness against jailbreaks, and adaptability to complex real-world scenarios \cite{zeng2024air,lin2024wildbench}. Safety evaluations are also cross-cutting :MCQs, Open-Response benchmarks, and Lab-style and Field-style Experiments. SafetyBench \cite{zhang2023safetybench}, for example, employs 11,435 MCQs in seven categories (e.g., bias, toxicity) to systematically test model adherence to ethical and safety standards in both English and Chinese. Similarly, SALAD-Bench \cite{li2024salad} proposed 4000 MCQs, and part of the larger dataset structured into a detailed hierarchy of 6 domains, 16 tasks, and 66 categories. BigBench \cite{srivastava2022beyond} has a subset of its tasks focused on safety evaluation in regards to toxicity, bias and truthfullness tat are MCQs. However, there has not been a focused MCQ-based safety evaluation benchmarks that take the nuances aspects of science, especially high consequential ones such as the the chemisty, biology, radiation, and nuclear (CBRN). We discuss one such effort on risks in chemistry in this work.
Open-response evaluations, such as those in DecodingTrust \cite{wang2024decodingtrust} and TrustLLM \cite{huang2024position}, examine nuanced safety aspects, including hallucinations, privacy violations, and machine ethics. TrustLLM evaluates LLMs across six dimensions, including fairness and safety, using over 30 datasets to identify critical safety gaps, while DecodingTrust introduces an eight-dimensional framework that probes issues like toxicity and ethical reasoning, with results published on a widely accessible leaderboard for transparency. However, such evaluations for scientific use cases are scarce~\cite{zeng2024air} and thus provide an opportunity to develop them in the future.

The safety red-teaming methodologies can be effectively interpreted along the lines of lab-style and field-style experiments. In lab-style red-teaming, researchers interact directly with LLMs and systematically introduce adversarial prompts to identify vulnerabilities such as biases, hallucinations, or ethical compliance issues. For example, \cite{buszydlik2023red} present a framework for red-teaming experiments on LLMs by generating numerical questions and puzzles to evaluate the models' performance on elementary calculations and algebraic tasks. This approach provides detailed feedback at each step, allowing iterative improvements and a deeper understanding of model limitations. In contrast, field-style red-teaming assesses LLMs by analyzing large-scale human interactions in real-world environments. This method captures diverse and unpredictable user inputs, offering insights into how models perform ``in the wild." For instance, \cite{hong2024curiositydriven} discuss automating red-teaming by training a separate red team LLM with reinforcement learning to generate test cases that maximize the chance of eliciting undesirable responses from the target LLM. This large-scale approach identifies practical weaknesses and vulnerabilities that may not surface in controlled lab settings, contributing to model robustness across varied real-world scenarios. By employing both lab-style and field-style red-teaming strategies, researchers can comprehensively evaluate and enhance the safety, reliability, and ethical performance of LLMs across different contexts. That said, significant work still needs to be done in designing such experiments for safety and trust scenarios in science.

\section{Establishing a methodology to evaluate LLMs as research assistants}


\begin{table*}
    \includegraphics[width=\textwidth]{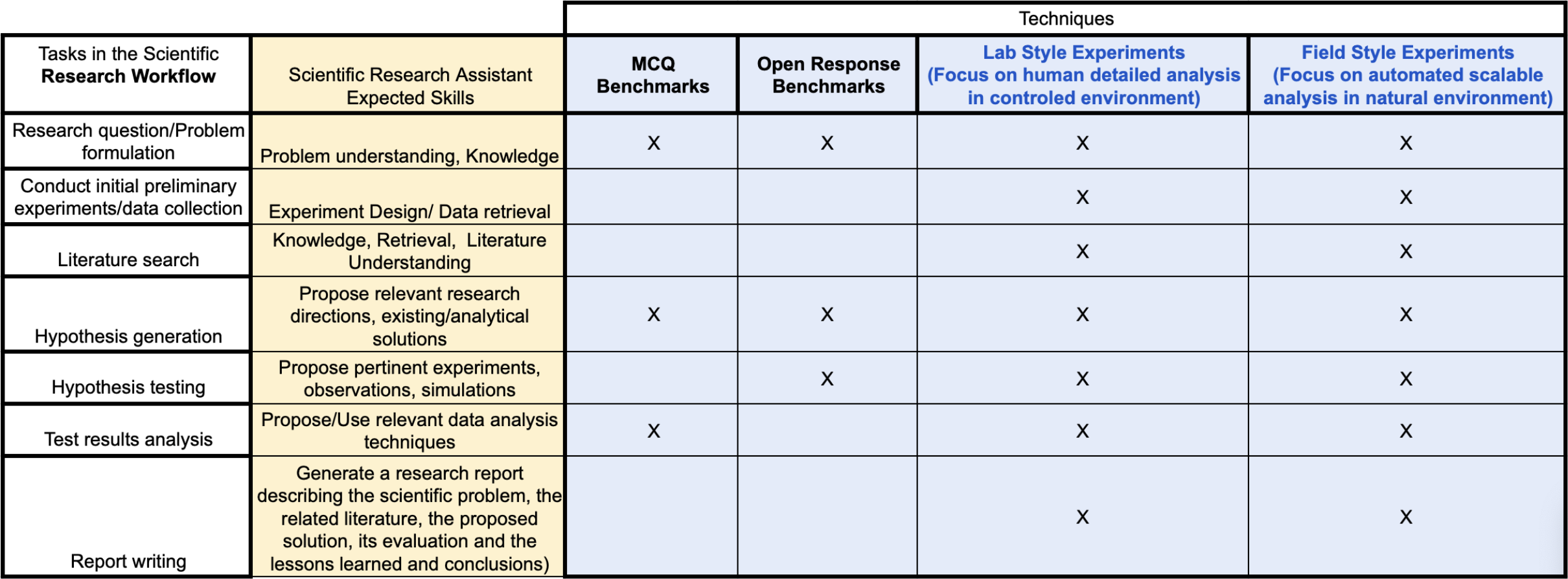}
    \caption{Skills evaluated by each of the evaluation techniques. Lab-Style Experiments focus on detailed analysis in controlled environment. Field-Style Experiments focus on analyzing researchers-LLMs interactions at scale in natural setting.}
    \label{tab:skills}
\end{table*}

The overarching objective of using LLMs as research assistants is to accelerate the research process.

In \autoref{tab:skills}, we present seven tasks commonly performed by researchers to solve a scientific problem: (i) posing and formulating a research question; (ii) if needed, conducting initial, preliminary experiments, observations, simulation, and/or database accesses to confirm the pertinence of the research question; (iii) performing literature search to identify related work; (iv) generating potentially multiple hypotheses (or research directions) to address the research question; (v) designing experiments, observations, and/or simulations to test the hypotheses; (vi) analyzing the resulting data to validate or invalidate the hypotheses; and (vii) writing a report about findings.  
Each of these seven tasks requires deep reasoning, contextual understanding, and iterative problem solving.
The scientific community needs confidence that LLMs are proficient in these tasks if they are to adopt them widely as research assistants. Thus a thorough and comprehensive evaluation methodology is required that can produce a rigorous assessment of LLM strengths and weaknesses in each task. 

As discussed in the related work section, most LLM evaluations take the form of MCQ benchmarks, which are a subset of broader Q\&A evaluations.
However, while these benchmarks serve a useful purpose in quickly assessing the breadth of knowledge of LLMs, that cannot, by their very nature, assess the depth of reasoning, contextual understanding, and iterative problem-solving required in the various steps of the scientific research process \cite{zhang2024massw, Pyzer-Knapp2022}. 

To address these limitations, it is critical to complement MCQ benchmarks with end-to-end evaluations in realistic contexts that assess both reasoning across multiple dimensions and the ability to plan and adapt across multi-step tasks. 
Therefore, we propose \textit{Evaluating AI models as scientific Research Assistants} (\textbf{EAIRA}), a structured evaluation methodology (\autoref{fig:multi_facet}) that combines four techniques:
two existing approaches that allow for quick and repeated assessment of the breadth of LLM abilities (MCQ Benchmarks and Open-Response Benchmarks) plus two new approaches for end-to-end evaluation of LLMs as scientific assistants (Lab-style Experiments and Field-style Experiments).

\textbf{MCQ benchmarks} serve to test foundational knowledge and domain-specific expertise across diverse scientific fields.  In addition to previously developed benchmarks, members of our team have developed three new MCQ benchmarks to address topic gaps in tasks for which MCQs are well suited.
1) Astro and 2) Climate are domain-specific benchmarks primarily generated through automated techniques to ensure the scalability of benchmark generation to serve new and evolving topics in science.
3) AI4S is a multi-domain ``AI for science'' MCQ benchmark, which integrates human question generation and validation with automated generation and validation methods to achieve a balanced, high-quality dataset that can be developed with moderate effort. By leveraging  both human and automated techniques, we can assess the strengths and weaknesses of automated methods deployed in Astro and Climate in order to guide future research in improving their generation.

\textbf{Open-response benchmarks} serve to test more detailed knowledge and to generate open-ended responses or code that assesses a model’s dynamic capability to handle unstructured, complex tasks, moving closer to reflecting the flexibility and adaptability required in real-world research scenarios while still facilitating a fast, automated evaluation.
For this, our team has previously developed two benchmarks: SciCode, which assesses the ability of LLMs to develop code which is highly dependent on the knowledge of the context of a scientific domain to correctly implement, and ALDbench, which assesses the ability of LLMs to describe methods for synthesizing materials using atomic layer deposition.

\textbf{Lab-style Experiments} perform evaluation by domain experts of AI models as they assist across all research tasks in real situations.  This expert-reviewed method provides a comprehensive assessment of the relevance, effectiveness, and iterative improvement of a model over time (e.g., 4-12 hours per attempt).  
This technique goes beyond open response by using a human proctor to guide an expert to iteratively interact with the model in order to assess multi-stage planning and response to results in the context of a scientific domain. Because of the interactivity, it provides very granular feedback about what models are or are not capable of performing giving a unique insight into the strengths and weaknesses of models and possible paths for improvement.
However, this capability comes at the cost of extensive effort by experts and proctors.

\textbf{Field-style Experiments} (inspired by, and adapting ``in-the-wild" evaluations \cite{Lin2024WildBenchBL,shen2024donowcharacterizingevaluating} for the scientific context) analyze automatically thousands of prompts, responses, and user behaviors during real-world interactions between researchers and AI assistants. This approach allows for large-scale capture and analysis of user needs, model strengths and weaknesses, and usage trends.
It differs from lab-style experiments in that:
1) experts are not guided by a proctor, 2) analysis of interactions and feedback is automated, and  3)  experiments can be conducted in the background as experts perform routine tasks (e.g., by capturing outputs of a site-wide LLM inference service or API proxy). These differences greatly improve the scalability of realistic end-to-end experiments but at the loss of the fine-grained granularity of the strengths and weaknesses detected by the lab-style approach.

In applying these four techniques, three key cross-cutting aspects must be considered: ethics, trust and safety; reliable uncertainty quantification; and scalable software infrastructure.
These three aspects ensure that LLMs are ``aligned'' with human values, produce and qualify results correctly under uncertainty; and can be evaluated efficiently at scale.
\textit{Trust and safety} evaluations must address ethical alignment, defend against jailbreak attempts, and adapt to complex real-world contexts. Multiple-choice strategies~\cite{zhang2023safetybench} can identify biases, toxicity, and compliance gaps, while open-response tasks~\cite{wang2024decodingtrust} can probe subtle issues such as hallucinations and machine ethics. Lab-style red-teaming~\cite{buszydlik2023red} systematically challenges the model with adversarial prompts in controlled settings to expose stepwise weaknesses, while field-style red-teaming \cite{hong2024curiositydriven}  tests the model’s robustness amid unpredictable real-world inputs, unveiling vulnerabilities that may remain hidden in laboratory conditions. \textit{Reliable uncertainty quantification} (UQ) is equally critical to establish trust in AI-driven scientific research assistants, as it systematically gauges model confidence and highlights potential inaccuracies \cite{lin2023generating, du2024haloscope, duan2024shifting}. UQ insights guide targeted refinements in MCQ, open-response, and lab- or field-style evaluations, ensuring that areas of high uncertainty are addressed in scientific and real-world contexts \cite{ye2024benchmarking, chen2024question}. Together, robust safety evaluations and UQ foster transparency, accountability, and trust, facilitating the responsible integration of LLMs into critical scientific research \cite{zeng2024air, achiam2023gpt}.
Finally, \textit{ the scalable software infrastructure} enables rapid evaluation to keep pace with rapid changes in the field of AI research.
The framework needs to incorporate distributed task parallelism \cite{babuji2019parsl} and fast inference capabilities \cite{kwonEfficientMemoryManagement2023}.

The following subsections present greater detail and results of the four techniques and three aspects of our evaluation methodology.

\label{sec:evaluation-team-activities}
\subsection{Domain-Specific MCQ Benchmarks}

Domain-specific benchmarks are crucial for evaluating LLMs in specialized fields, as they address the limitations of general benchmarks that often fail to capture the complexities of domain-specific tasks. Without tailored benchmarks \cite{ting2024astromlab}, LLMs risk over training on well-established datasets, leading to inflated performance that does not translate to real-world applicability.
These benchmarks are essential for guiding targeted improvements, ensuring that models meet the specific demands of scientific research, and providing a baseline to understand their strengths and weaknesses. By capturing the nuanced challenges of individual domains, such benchmarks foster the effective and ethical deployment of LLMs, enabling their potential to accelerate discovery and innovation across disciplines. We now discuss the domain-specific benchmarking efforts that we conducted in Astronomy and Climate modeling.

\subsubsection{\bf Astronomy} 
The Astronomy Benchmark \cite{ting2024astromlab} assesses LLM performance in a manner that reflects the interdisciplinary nature of astronomy, testing both factual recall and the ability to connect insights across subfields. This benchmark was generated automatically using an LLM to compose MCQs from astronomy papers. To assemble a rich repository of scientific knowledge, we leveraged the Annual Review of Astronomy and Astrophysics (ARAA), a selective review journal renowned for its comprehensive overviews authored by leading experts. 

The Nougat optical character recognition (OCR) tool was used to transcribe 885 ARAA articles over the years 1963 to 2023. Each transcribed paper was processed using Gemini-1.5-Pro, a long-context LLM capable of handling up to one million tokens, to generate five multiple-choice questions (MCQs) per paper. 
The questions were designed to be specific, yet independent of the article sections, with generalized answers and balanced options to avoid bias. This process yielded a total of 4425 MCQs covering diverse topics such as quasar density decline at high redshifts and subgrid feedback model calibration in simulations. 

\begin{tcolorbox}[colback=blue!10, colframe=blue!40!black, title=Sample question from Astronomy benchmark dataset, left=2pt, right=2pt, top=1pt, bottom=1pt]
\textbf{How does the presence of stellar companions influence the formation and detection of exoplanets?}
\small
\begin{enumerate}
    \item[(A)] Stellar companions can dilute transit signals, potentially leading to misclassification of planets and inaccurate parameter estimations. Additionally, their gravitational influence can suppress planet formation in close binary systems.
    \item[(B)] Stellar companions provide additional sources of gravitational perturbations, enhancing planet formation by promoting planetesimal accretion and facilitating the formation of gas giants.
    \item[(C)] Stellar companions contribute to the metallicity enrichment of planetary systems, leading to the formation of more massive and diverse planets, including super-Earths and hot Jupiters.
    \item[(D)] Stellar companions act as gravitational lenses, increasing the detectability of exoplanets through microlensing events and enabling the discovery of planets at greater distances from their host stars.
\end{enumerate}
\end{tcolorbox}

The Astronomy Benchmark has been then used to assess both the accuracy and computational cost of dozens of different closed and open LLM variants.
This study revealed disparities in LLM performance across general-purpose and specialized tasks that highlight significant performance gaps and performance-to-cost ratios.
While frontier models like 
Claude-3.5-Sonnet excelled in the Astronomy Benchmark with an accuracy of 85.0\%, outperforming GPT-4o (80.4\%) and Gemini-1.5-Pro (77.6\%), these differences are less evident in general-purpose benchmarks such as MMLU~\cite{open-llm-leaderboard}. 
A study of how Astronomy Benchmark performance varied with compute costs showed that, roughly speaking, each 3.5 percentage points increased accuracy was associated with 10-fold increase in price, within most given series of models such as GPT, Gemini or Claude. An analysis of the cost per 0.1 M tokens showed that the cost for a desired performance can vary by more than three orders of magnitude across different models: see Figure~2 in \cite{ting2024astromlab}.

The study also showed that open-weight models, though improving, lag behind proprietary ones, with older versions underperforming by as much as 30\% on specialized tasks. Performance also varies significantly between English-focused and non-English-focused models, with the latter struggling in areas like theoretical astronomy and advanced instrumentation, reflecting gaps in training datasets. However, recent astronomy subfields, such as post-1990 advancements, exhibit narrower performance gaps which may be
due to model's ability to handling historical context or older scientific consensus.
These findings emphasize the importance of domain-specific benchmarks to assess not only performance in specialized tasks but also the performance-to-cost ratio crucial for user adoption in assisting with scientific research. This work also shows that performance varies across sub-fields.


\subsubsection{\bf Climate}
Climate and weather forecasting presents multifaceted challenges that demand interdisciplinary knowledge and reasoning, making it an ideal testbed for evaluating LLM capabilities.
However, existing benchmarks for climate science are limited \cite{lacombe2023climatex, vaid-etal-2022-towards, thulke2024climategpt}. 
Thus we developed a Climate Benchmark, a set of MCQs focused on the urgent and complex domain of climate science.   
As the manual creation of such MCQs is labor intensive and climate scientists are already overcommitted, we employed automated methods to develop the Climate Benchmark. 
We adopted as our source material the Intergovernmental Panel on Climate Change (IPCC) reports \cite{solomon2007ipcc}, which are authoritative assessments synthesizing the latest scientific research on climate change, its impacts and potential solutions. These reports serve as a foundation for informed decision-making and international negotiations on climate action, highlighting the urgency of addressing the complex and interconnected challenges posed by changing climate. The reports are typically extensive, often exceeding 1000 pages, with each chapter and section addressing specific topics related to climate change.
To generate questions on the various topics discussed in the report, we parsed the PDFs section-wise, using Nougat as was done for the Astronomy Benchmark. We then employed OpenAI's GPT4 to create one multiple-choice question, consisting of one correct answer and four distractors, for each section, with the prompt designed to create an MCQ based on the provided scientific text, ensuring that the question evaluates a broader understanding of climate science principles without referencing the specific report.
This process resulted in a total of 752 questions on diverse topics, ranging from factors influencing vulnerability to climate change to primary strategies for risk reduction. This systematic approach ensured comprehensive coverage and alignment with the content of the IPCC reports. 
 
We employed the Climate Benchmark to assess the performance of multiple LLMs, including GPT-4o, Llama~3.8B, and Phi~4, on specialized climate science tasks. Among these models, GPT-4o performed the best, achieving an accuracy of 87.34\%, demonstrating its effectiveness in handling complex climate-related tasks. GPT-4o was followed by Llama~3.8, which achieved an accuracy of 78.48\%, and Phi~4, which scored 54.43\%. This performance disparity highlights the need for continued refinement and optimization of models to bridge the gap in specialized applications.

The development of the Climate Benchmark provided useful insights into the creation and evaluation of MCQ datasets for scientific applications. The climate-focused MCQs, derived from IPCC reports, were designed to assess knowledge recall and decision-making, emphasizing accurate understanding of scientific concepts. While primarily testing factual knowledge, these questions establish a strong foundation for expanding into tasks that require more complex reasoning and application in climate science. However, the automatic generation of MCQs sometimes produced semantically similar questions that differed only slightly in phrasing or structure while testing the same core information. This observation highlights the need for robust evaluation mechanisms to eliminate redundancy and ensure diversity within the dataset. While automatic MCQ generation greatly accelerates the creation of benchmark questions, it cannot replace a rigorous evaluation process. A combination of LLM-based evaluators and human oversight is crucial for maintaining the benchmark's quality, relevance, and accuracy, ensuring that it meets the standards required for research-focused benchmarks.

\subsection{Scalable AI4S MCQ Benchmark}

\begin{figure*}[htb] 
    \centering
    \hrule
    \includegraphics[width=\textwidth, clip]{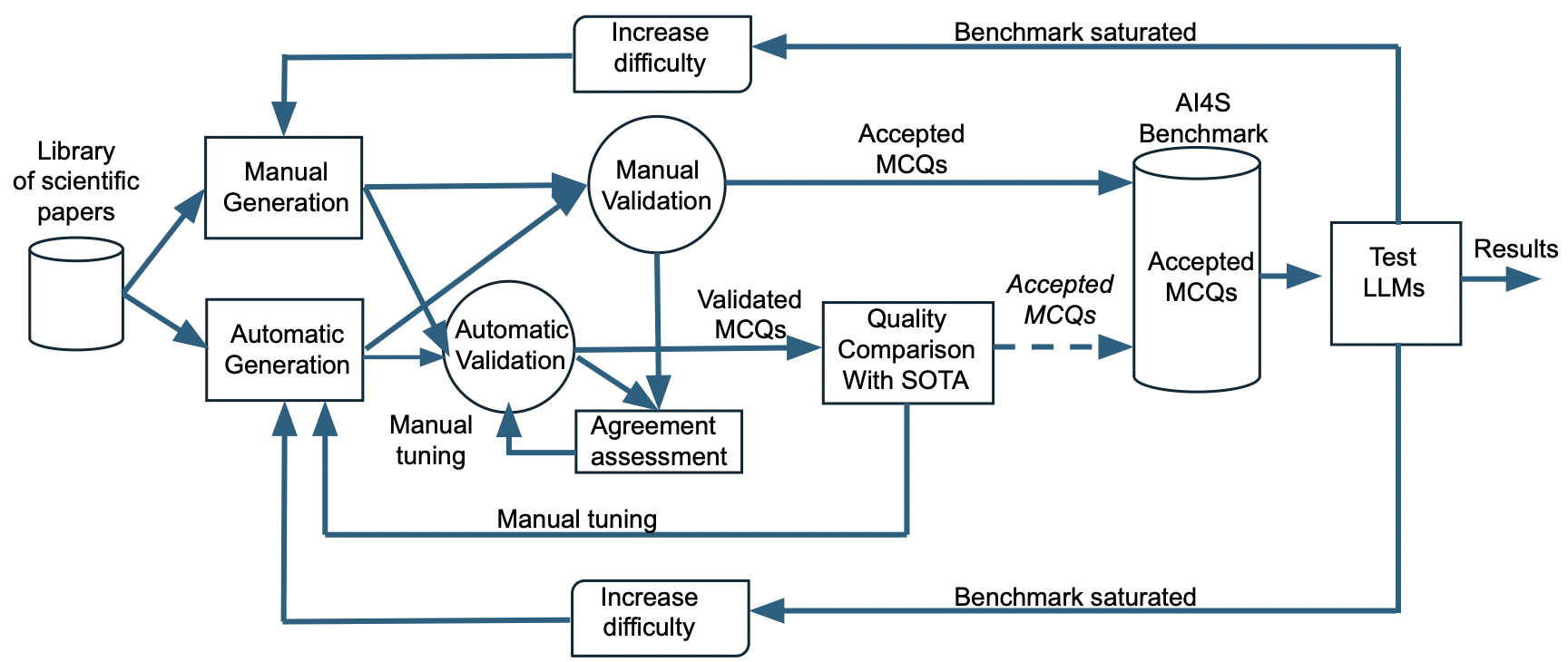} 
    \vspace{-2mm}
    \hrule
    
    \caption{The AGIL approach to generate scalable MCQ benchmarks. The current version of the AI4S benchmark contains only manually accepted MCQs. The AGIL approach  enables the integration of automatically accepted MCQs after the validation of their difficulty and quality.} 
    \label{fig:AGIL}
\end{figure*}

The overarching goal of the AI4S benchmark is to evaluate the knowledge extension of LLMs across many different scientific domains.
In that respect, it is similar to GPQA \cite{rein2023gpqa}. However, AI4S design focuses on the quality and scalability of MCQ generation and validation. While GPQA has only 448 MCQs, our objective is to generate and validate thousands of MCQs and continuously add MCQs as LLMs progress in their capabilities.

Current MCQ benchmarks, including GPQA, have two main limitations: 1) they are quickly saturated because of the fast progress of LLMs. 2) there is a high risk of contamination (benchmark included in the training sets) if the MCQs are made public. These two limitations arise from the static aspect of the current benchmarks. They are developed once and do not evolve. The current practice is to develop new versions \cite{GSM1k} when the initial benchmark is saturated \cite{gsm8k} and to open only a portion of the benchmark MCQs to avoid contamination. 

To address the two limitations, we explore a novel approach to develop scalable generation and validation of MCQ benchmarks for the evaluation of LLMs capabilities: Automatic continuous Generation and validation of Increasingly Large MCQ benchmarks: AGIL.

Building on the experience gained from the Astronomy and Climate domain-specific benchmarks, we are developing an AGIL process by which we combine manual and automated methods to generate and validate MCQs. 

We present here our initial finding towards the creation of a 1000-MCQ AI4S benchmark that spans five scientific domains: Computer Science, Astrophysics, Climate, Physics, and Chemistry. Our goal with this first study is to assess key aspects of the AGIL approach: validity of the difficulty level, quality of the generated MCQs compared to GPQA ones, quality of the automatically generated MCQs compared to manually generated ones, and quality of the automatically validated MCQs.

In the AGIL approach (Figure \ref{fig:AGIL}), manual MCQ generation and validation by experts is critical to provide high-quality, domain-specific MCQs that serve as a gold standard for evaluating LLM performance and developing automatic MCQ generation and validation workflows. The goal of automated generation and validation (LLM as a judge technique \cite{LLM-as-judge}) by LLMs is to address the scalability issues of manual generation and validation while keeping their quality levels. 

\begin{table*}[htbp]
\caption{Performance metrics for Llama~3 models across various AI4S benchmark levels.
}
\label{tab:performance_metrics}
\begin{tabularx}{\textwidth}{Xlrcc}
\hline
\textbf{Model} & \textbf{Task} & \textbf{nsamples} & \textbf{acc (stderr)} & \textbf{acc\_norm (stderr)} \\
\hline
\textbf{Llama-3-8B}   & accepted        & 254 & 0.2008 (±0.0252) & 0.2717 (±0.0280) \\
\ \ \ \textbf{Llama-3-8B}    & accepted\_easy   & 81  & 0.2222 (±0.0465) & 0.3210 (±0.0522) \\
\ \ \ \textbf{Llama-3-8B}    & accepted\_medium & 116 & 0.1983 (±0.0372) & 0.2672 (±0.0413) \\
\ \ \ \textbf{Llama-3-8B}    & accepted\_hard   & 57  & 0.1404 (±0.0464) & 0.1930 (±0.0527) \\
\textbf{Llama-3.1-8B}  & accepted        & 254 & 0.1969 (±0.0250) & 0.2638 (±0.0277) \\
\textbf{Llama-3-70B}   & accepted        & 254 & 0.2598 (±0.0276) & 0.3701 (±0.0304) \\
\textbf{Llama-3.1-70B} & accepted        & 254 & 0.2520 (±0.0273) & 0.3386 (±0.0298) \\
\hline
\end{tabularx}
\end{table*}

For the manual generation and validation of MCQs we organized hackathons which engaged 140 domain experts (PhDs and other research staff) \footnote[1]{\label{note1}the exercises in this paper were done with volunteer Argonne employees, who understood that the goal of this effort is to identify opportunities to improve models.  While the research team offered a rubric for notes and observations, the participants were free to use whatever rubric they preferred.  The following is the general approach that was used.}. We offered the participants to generate MCQs from scientific papers of their choosing, including their own. The manually generated MCQs were crafted by using a purpose-built authoring interface (Appendix, \autoref{fig:questionauthoring}) that allowed contributors to test their questions on smaller models like Llama~3 before submission so as to ensure a baseline difficulty threshold. Manual validation used another specifically created form (see Appendix, \autoref{fig:questionreviewing}). The tool used for the MCQ generation and validation is available on github (https://github.com/auroraGPT-ANL/questions-ui)

Automated MCQ generation leveraged LLMs such as GPT-4 with domain-specific prompts to create MCQs from scientific papers across fields such as climate science, physics, and chemistry, with validation guided by the LLM-as-a-Judge technique. Both the prompts used for automatic MCQ generation and validation, as well as the rubrics for validation and reviews, were informed by the manual generation and validation, as well as by the experience gained during the domain-specific benchmark development discussed earlier. 

This process generated 980 MCQs (720 manually and 260 automatically). Of 588 total manually generated reviews, 317 MCQs have been assessed so far, of which 254 were accepted. Acceptance was subject to the following criteria: appropriate difficulty, relevant, complete and correct answers and distractors, controversial answers, mathematic requirements, as well as relevant skills and domain selection. The small percentage of accepted MCQs 25\% illustrates the difficulty of generating and validating high-quality scientific MCQs.

In addition, domain experts categorized the accepted MCQs into easy, medium, and hard levels, capturing a spectrum of difficulty that mirrors real-world scientific challenges. This multi-level structure enables the AI4S benchmark
to evaluate both the foundational and advanced capabilities of LLMs, offering an assessment of their strengths and limitations. 

To evaluate the merits of our AGIL approach, we performed several tests to evaluate the quality of (i) the level classification, (ii) the AI4S benchmark compared to GPQA, (iii) the automatically generated MCQs, and (iv) the automatic MCQ validation.

For all tests, we used the STaR framework (see \autoref{sec:framework}) with five shots.
The first row of the table \autoref{tab:performance_metrics} shows the overall performance of Llama~3~8B on the 254 MCQs. The resulting accuracy of 20\% correspond to a random guess. 
The next three rows show results for MCQs grouped by their human-identified levels of difficulty. We see that the Llama~3~8B results are consistent with human-identified difficulty levels, with Llama~3~8B achieving the best score on easy MCQs (22\%) and significantly below random performance on hard MCQs (14\%). These results validate the quality of the level classification.
The next three rows of \autoref{tab:performance_metrics} show the performance of other Llama-3 models for the AI4S accepted MCQs.
We observe that Llama~3-70B performs less well on all AI4S-accepted MCQs (26\% of questions answered correctly on all difficulty levels) than on GPQA (49\% of questions answered correctly). Note that while GPQA has one correct answer and three distractors, meaning 25\% accuracy for random responses, AI4S MCQs have four distractors and thus only 20\% random response accuracy. We conclude that AI4S is a more challenging LLM benchmark than GPQA. 

To obtain a finer quality comparison between AI4S and GPQA MCQs, we use automatic MCQ validation to quantify the quality of every MCQ on the first seven criteria (N/A value for the eighth criterion on GPQA). \autoref{tab:mean_sd_scores} shows the scores of the two benchmarks on the seven criteria. On average, AI4S MCQs (average of 4.55) reach the same quality level as the GPQA MCQs (average of 4.45). 

\begin{table}[htbp]
\centering
\caption{Mean and standard deviation (SD) scores for AI4S and GPQA across various criteria.}
\label{tab:mean_sd_scores}
\begin{tabular}{lcc}
\hline
\textbf{Item} & \textbf{AI4S Mean (SD)} & \textbf{GPQA Mean (SD)} \\
\hline
Appropriate & 3.68 (0.72) & 4.28 (0.58) \\
Complete & 4.52 (0.87) & 4.42 (0.75) \\
Controversial & 4.97 (0.19) & 5.00 (0.04) \\
Correct & 4.60 (1.36) & 4.21 (1.82) \\
Domains & 4.68 (0.73) & 4.97 (0.25) \\
Mathematic & 4.81 (0.95) & 3.50 (2.30) \\
Relevant & 4.64 (0.56) & 4.81 (0.42) \\
Skills & 3.97 (0.70) & N/A \\
\hline
\end{tabular}
\end{table}

Overall, these results validate the quality of AI4S compared to GPQA.



We used the acceptance criteria to compare the quality of the manually and automatically generated MCQs. This comparison reveals that the quality of manually validated, automatically generated questions is on par with manually validated/generated ones. Overall these results show that automatic MCQ generation can be used to generate MCQs with a manual verification step.  

To evaluate the quality of automatic MCQ validation, we compare it with manual validation. We used Mistral Large~2 as the judge, prompting it to evaluate each MCQs 
on a scale of 1--5 for each of eight criteria and also asking it to provide concise rationale for each score. All criteria were evaluated in a single prompt. We expanded each criteria specification to define each level on the scale for all criteria explicitly.
The prompts used for the judge are in the Appendix, \autoref{fig:ai4s-judge-prompt}.
We observe the accuracy of a model trained on these judgements to predict whether a question will be accepted/rejected to be 72\%. For comparison, of the 144 questions with multiple reviews, only 61\% of reviewers agreed on acceptance.

During the development of the AI4S benchmark using our proposed AGIL approach, several key lessons emerged.
Generating high-quality scientific MCQs manually is a challenging task for multiple reasons: crafting questions with distinguishable levels of difficulty (undergraduates, PhD students, postdocs, and staff)
is non-trivial, and creating distractors that are plausible but not overly confusing requires significant effort and precision. Validation of these questions proved even more demanding, as finding appropriate reviewers for difficult and specialized topics can be challenging, and ideally, each question requires validation by three experts to ensure reliability. 
Our goal is to continue the development of the AGIL approach to address this difficulty issue and to continuously generate AI4S MCQs from the large pool of available scientific papers. We will release the benchmark using a sliding-window approach, keeping a significant portion of newly generated MCQs (e.g. 70\%) not public. 


\subsection{Open Response Benchmarks}
Next we discuss the open-ended benchmarks. 
These are essential for evaluating the reasoning, creativity, and problem-solving abilities of LLMs, particularly in scientific domains. Unlike MCQs, which primarily test factual recall or single-step reasoning, open-ended tasks engage models with complex, real-world problems that require multi-step reasoning, synthesis of knowledge across domains, and adaptive problem-solving. For instance, while an MCQ might test a model's recall of a specific scientific fact, an open-ended task could require the model to design an experiment, analyze data, or propose solutions to unsolved research questions. This format aligns better with the exploratory nature of scientific inquiry, offering a more comprehensive assessment of a model's capabilities. However, existing open-ended benchmarks often fall short in capturing the depth and realism needed for scientific evaluations. Many rely on synthetic tasks that fail to reflect the intricacies of real-world scientific challenges, such as multi-disciplinary reasoning or generating accurate code for practical applications. 

While open responses questions are versatile in capability, their
evaluation is more complicated compared to MCQs.
Different assessment approaches are applied, depending on the task at hand.  
A first class of \textit{statistical scorer} approaches looks at co-occurrence of n-grams (sequences of letters or words) in LLM outputs vs.\ supplied ground truth responses. 
In this context, widely used scores are BLEU (BiLingual Evaluation Understudy) compares LLM outputs against ground truths. It calculates the precision for each matching n-gram (n consecutive words) between an LLM output and expected output. ROUGE (Recall-Oriented Understudy for Gisting Evaluation) evaluates text summaries and computes recall by comparing the overlap of n-grams between LLM outputs and expected outputs. It determines the proportion (0–1) of n-grams in the reference present in the LLM output. METEOR (Metric for Evaluation of Translation with Explicit Ordering) calculates scores by assessing both precision (n-gram matches) and recall (n-gram overlaps) and leverages linguistic databases to account for synonyms. Statistical scorers do not take any semantics and reasoning capabilities into account. 

A second class are \textit{embedding approaches} that seek to compare the semantics of LLM responses and reference answers. Here, some widely used scores include BERTScore, that relies on pre-trained language models (e.g., BERT) and computes the cosine similarity between the contextual embeddings of the LLM responses and references. These similarities are then aggregated to produce a final score. Other tools such as CheckEmbed \cite{besta2024checkembed} can be used to compare the semantics of LLM responses and reference answers. 
The third and most recent class is the  \textit{LLM-as-a-judge methods} which tackle the problem of evaluating LLM open responses when no reference answer is available. This approach currently has two variations. In the ``Pairwise comparison" version, an LLM judge is presented with a question and two answers, and tasked to determine which one is better or declare a tie. In the ``Single answer grading" version, an LLM judge is asked to directly assign a score to a single answer. In principle, LLM-as-a-judge can offer several key benefits: consistency, scalability, and explainability. However, the approach also has limitations: position bias (first answer better in ``Pairwise comparison"), verbosity bias (longer answer better), self-enhancement bias (self-generated answer better) and limited capability to grade math and reasoning questions \cite{LLM-as-judge}. Moreover, the reliability of such evaluations is still the subject of research.

We now discuss two open-ended benchmarks, one for scientific code generation and another for atomic layer deposition in Material Science.

\subsubsection{\bf SciCode - Scientific Code Generation Benchmark}
The SciCode Benchmark is a set of manually curated coding problems designed to assess LLM capabilities for solving complex scientific coding problems across diverse domains. 
By providing tasks that reflect real-world challenges and require multi-step reasoning, SciCode allows models to be tested in contexts that align closely with the demands of scientific research~\cite{tian2024scicode}.
SciCode includes problems 
across a range of scientific domains, including computational mechanics, quantum information, quantum chemistry, ecology, and molecular modeling. It consists of 80 main problems, decomposed into 338 intermediate steps, enabling a structured approach to assessing model capabilities. Solving each individual problem requires that an LLM implement multiple Python functions corresponding to subproblems and integrate those functions into a cohesive solution: see \autoref{fig:scicode_example}. Each problem is accompanied by a gold-standard solution and multiple test cases so as to permit robust and reliable automatic evaluation. 
\begin{figure*} 
    \centering
    \includegraphics[width=\textwidth,trim=0 6mm 6mm 10mm, clip]{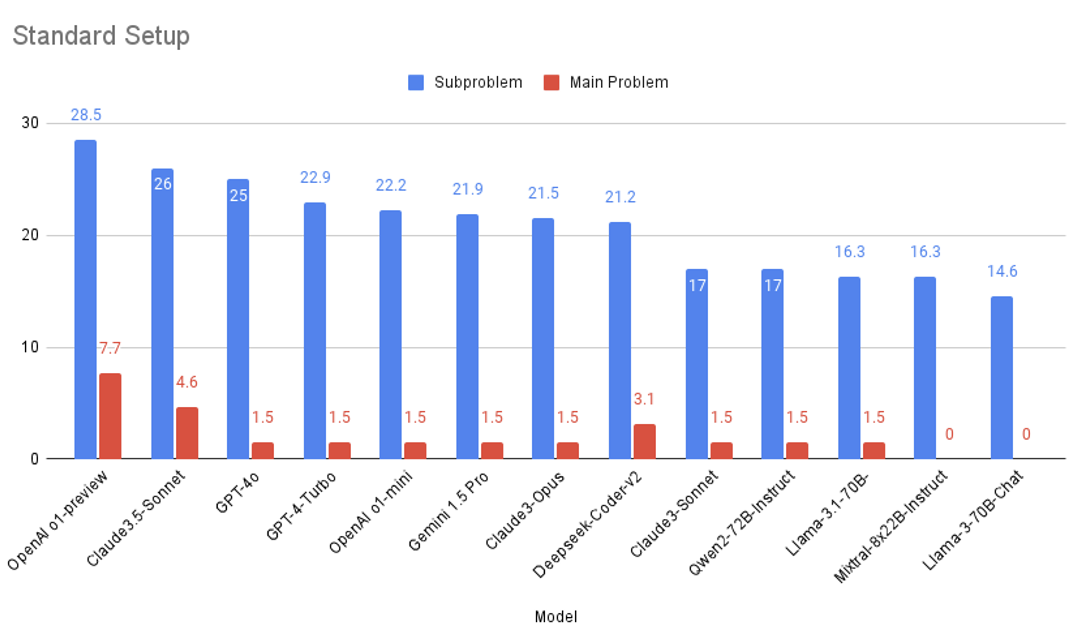} 
    \caption{The performance of various LLMs on SciCode problems. 
    This histogram displays the accuracy (vertical axis, 0\% to 100\%) of various state-of-the-art LLMs (listed on the horizontal axis) 
    in solving both main problems (red) and their associated subproblems (blue) within SciCode. 
    To solve a main problem, LLMs must implement one Python function per subproblem 
    and integrate them into a comprehensive solution. SciCode provides gold-standard solutions and multiple test cases for reliable automatic evaluation. 
    These consistently poor results highlight the need for LLMs that incorporate scientific knowledge and advanced reasoning to better assist researchers.  
    }
    \label{fig:scicode}
\end{figure*}


Each SciCode problem is meticulously annotated and verified by at least two senior researchers to ensure accuracy, and is drawn from real-world research tasks, maintaining relevance to practical applications. 
Problems are curated to avoid overlap with publicly available datasets and thus to test the deep scientific knowledge and analytical skills of LLMs by requiring the decomposition and integration of complex problems into comprehensive solutions. Additionally, SciCode 
allows for flexible evaluation of model capabilities in varied setups, enabling adjustments like providing background information or conditioning on previous solutions. 

The SciCode Benchmark is configured to assess LLM capabilities to solve SciCode problems by using zero-shot prompts, maintaining a general approach while creating distinct prompts for various evaluation setups to guide the model on the tasks, as 
described in detail in \cite{tian2024scicode}. 
The prompts remain consistent across models and fields, incorporating instructions for the main and sub-problems, as well as code from previous subproblems. We evaluated the coding capabilities of several state-of-the-art LLMs using the SciCode benchmark, focusing on three key aspects to assess their performance. First, the \textit{Impact of Scientific Background} was analyzed by testing models in two modes: without background, to evaluate inherent scientific knowledge and reasoning, and with background, to focus on coding and instruction-following capabilities. The results showed significant performance improvements with background information, highlighting the limitations of current LLMs in scientific reasoning. Second, the comparison between \textit{Gold vs. Generated Solutions} revealed insights into the challenges of realistic evaluations. While gold solutions accurately address each problem, generated solutions introduce error accumulation, creating a more practical and demanding evaluation scenario. Lastly, the assessment of \textit{Main vs.\ Subproblems} provided a nuanced understanding of model performance. A main problem was considered solved only when all subproblem solutions and the integrated result were accurate. Additionally, SciCode’s design allows independent evaluation of subproblems, enabling precise analysis of models’ reasoning and coding abilities across discrete tasks. These evaluation dimensions underscore the benchmark's rigor in testing LLMs for real-world scientific applications.

We summarize the findings of our studies using several state-of-the-art models in \autoref{fig:scicode}. These results show that SciCode is a difficult benchmark for current LLMs. Consistent with our observations on proprietary models, open-weight LLMs under test also showed their lack of capabilities in solving any main problem despite being able to solve a number of sub-problems correctly.

The SciCode project provides insights into the challenges of evaluating LLMs in scientific coding tasks, highlighting significant gaps in current capabilities. Despite recent advancements, state-of-the-art models like OpenAI’s o1-preview and Claude3.5-Sonnet solve only a small fraction (7.7\%) of the main problems, underscoring the disparity between existing LLMs and the deep scientific reasoning required for real-world research. SciCode is designed to address this gap by focusing on real-world, research-level problems across diverse natural science fields, including mathematics, physics, chemistry, and biology. Sourced from peer-reviewed work, these problems test LLMs' ability to generalize to less familiar domains. By decomposing problems into subproblems with detailed annotations, SciCode rigorously evaluates models’ coding, reasoning, and knowledge integration capabilities. While providing scientific background information improves model performance, the persistent struggle of LLMs with these tasks emphasizes their current limitations in handling complex scientific challenges. The project highlights the importance of high-quality data, domain-specific validation, and carefully curated problems to advance the development of AI tools for scientific research. The findings indicate substantial progress is needed to enhance scientific reasoning and background knowledge integration in LLMs to enable their effective application in real-world scenarios.
\subsubsection{\bf ALDbench - Materials Synthesis Benchmark}


An area that lacked relevant benchmarks is materials synthesis.
This is particularly important for potential applications
of LLMs in automated materials discovery or as AI research
assistants. LLMs underpinning such capabilities need to exhibit both the ability to reason about specific processes (for instance to avoid unsafe conditions or transfer ideas across reactors and process conditions) and have a robust understanding of the literature (to build on existing process knowledge and avoid known dead ends).

As such capabilities appear hard to evaluate by using either MCQs or the statistical scorer or embedding approaches described earlier.
we developed a new open-response benchmark ALDbench on materials synthesis, and in particular on a synthesis technique called 
atomic layer deposition \cite{aldbench}. Here we targeted a range of difficulty spanning from graduate level to PhD-level domain expert current with the state of the art. A model's ability to perform at a domain expert level is paramount whenever models are expected to assist in decision making processes that involve costly experiments.
Beyond its applied interest in areas such as energy and microelectronics \cite{AlvaroALD}, this domain brings together multiple
topics that are commonplace in chemistry-driven synthesis, including metal-organic and inorganic molecules, gas-surface kinetics and heterogeneous reactions, and gas phase transport. Evaluating LLM capabilities in this field can provide insights with wide applicability to other material synthesis techniques.

To compile the benchmark, we asked six PhD-level human experts to generate ``questions that a researcher or a graduate student who is not familiar with a specific process/application would ask an AI assistant." 
The curated questions could be grouped into four categories: 1) \emph{how to grow}, where the query is about material
synthesis; 2) \emph{specific questions about ALD processes}, comprising more in-depth queries about a process or material;  3) \emph{general ALD knowledge}, with questions about the synthesis
technique; and 4) \emph{applications}.

The human experts were then asked to grade the questions using a scale of 1 to 5 on two criteria with the following rubrics similar to the AI4S benchmark: (1) \emph{Difficulty:} 1--Easy, early graduate; 5--Hard, top expert; (2) \emph{Specificity:} 1--General; 5--Specific, quantitative.

Each response is then graded using four criteria with the following rubrics: (1) \emph{Overall quality:} 1--Very low quality; 5--Excellent; (2) \emph{Specificity:} 1--Too broad; 5--Targeted; (3) \emph{Relevance:} 1--Irrelevant fluff; 5--Relevant answer; (4) \emph{Accuracy:} 1--All made up; 5--All correct.   
The use of multiple criteria allowed us to probe aspects of
the generation process, such as relevance or specificity of the
response, that are not easily captured by benchmarks focused
on accuracy.

\begin{figure}[t]
\centering
    \includegraphics[width=.9\columnwidth]{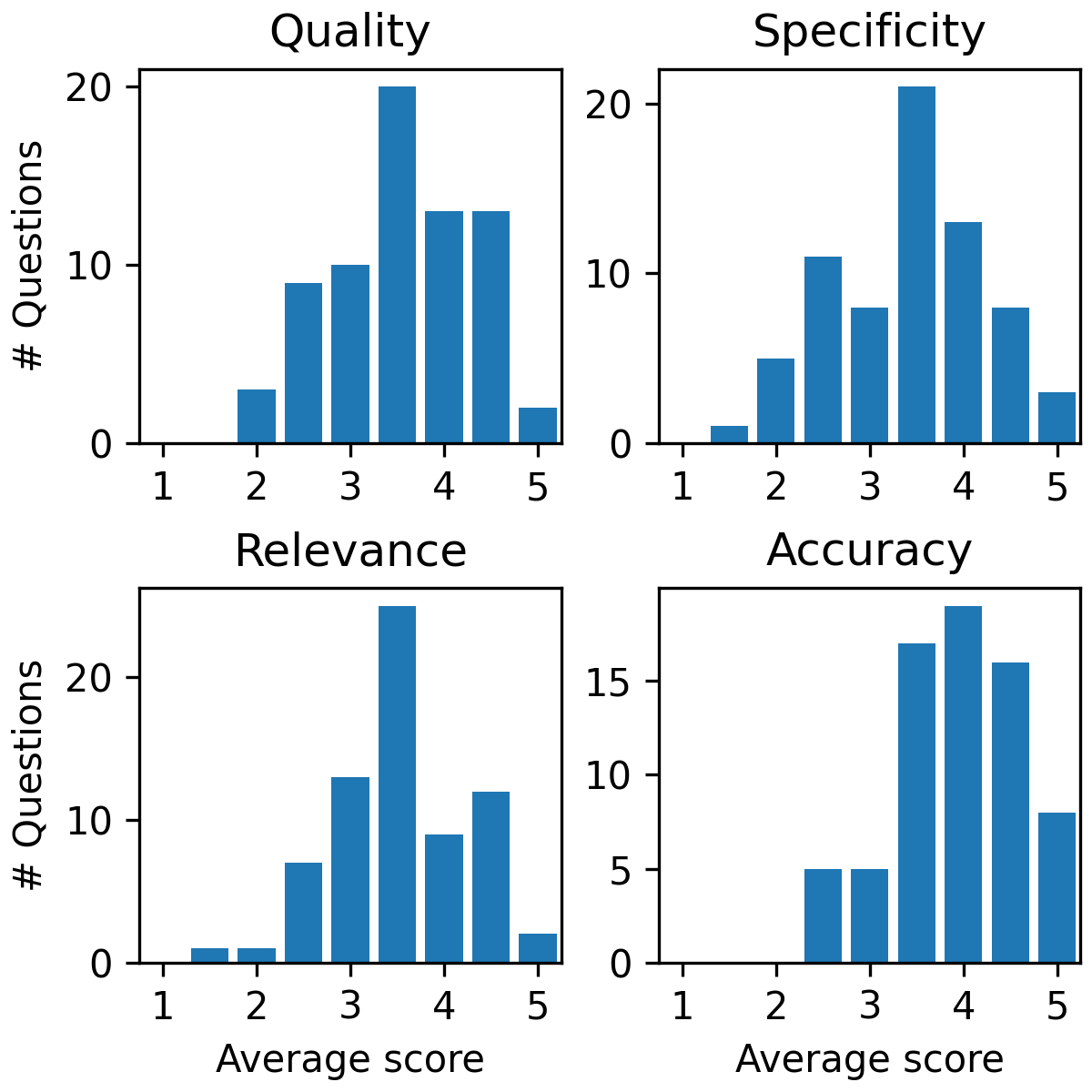}
    \caption{Distribution of the mean scores of GPT-4o
    responses to all questions in the ALDbench benchmark.}
    \label{fig:aldbench}
\end{figure}

We ran this benchmark using an instance of OpenAI’s GPT-4o, with seven PhD-level human experts reviewing model responses.
Details are in the ALDbench paper \cite{aldbench}.
The model responses received a composite quality score of 3.7, consistent with a passing grade. However, 36\% of the questions received at least one below average score. When
we carried out an in-depth analysis of the responses we
identified at least five instances of hallucination. In \autoref{fig:aldbench} we show the
distribution of mean scores for all the questions in the benchmark
and the four criteria evaluated by the human experts.

We also explored statistical
correlations between the difficulty and specificity of
each question and the human expert scores for each evaluation criteria. For each (question, response) pair we computed p-values using the Fisher
exact test to evaluate the statistical significance of the
correlation.  We found statistically significant correlations
between question difficulty and response quality ($p_0$ = 0.033), question difficulty 
and relevance ($p_0$ = 0.016), and question specificity
and response accuracy ($p_0$ = 0.007). In all three cases, higher difficulty or specificity correlated
with lower scores.
These results emphasize the need to evaluate LLMs across multiple criteria beyond traditional metrics of difficulty and accuracy.

Our results show that highly targeted, open-response benchmarks can provide
information about LLM performance in scientific domains that is complementary to MCQs or natural language processing benchmarks. The methodology developed in this work allowed us to probe in depth model performance in a specific domain. With the aid of a small team of PhD-level experts we were able to identify instances of hallucinations and explore model responses in a level of detail that it is hard to accomplish using automatic evaluation methods. The extension of this approach to other domains, such as energy storage or microelectronics, is trivial. Moreover, as a byproduct of this
effort, we collected a small dataset of questions and human rated responses across four different evaluation criteria. As we explore other domains we can use this data to train or validate automatic question evaluation approaches for open-ended benchmarks. 
\subsection{End-to-End Evaluations}
Although MCQ benchmarks are effective in testing factual recall and reasoning within constrained formats, and open-ended benchmarks gauge the generation of detailed and flexible responses, these methods do not capture the iterative and complexity of scientific problem solving. 
End-to-end evaluations attempts to address this gap by assessing, in real situations, the models responses for assisting researchers in solving scientific problems.
We propose two novel types of end-to-end methods in the context of scientific research: Lab-style and field-style experiments.  


\subsubsection{\bf Lab-style experiments}

\begin{figure*}[htb] 
    \centering
    \hrule
    \includegraphics[width=\textwidth, clip]{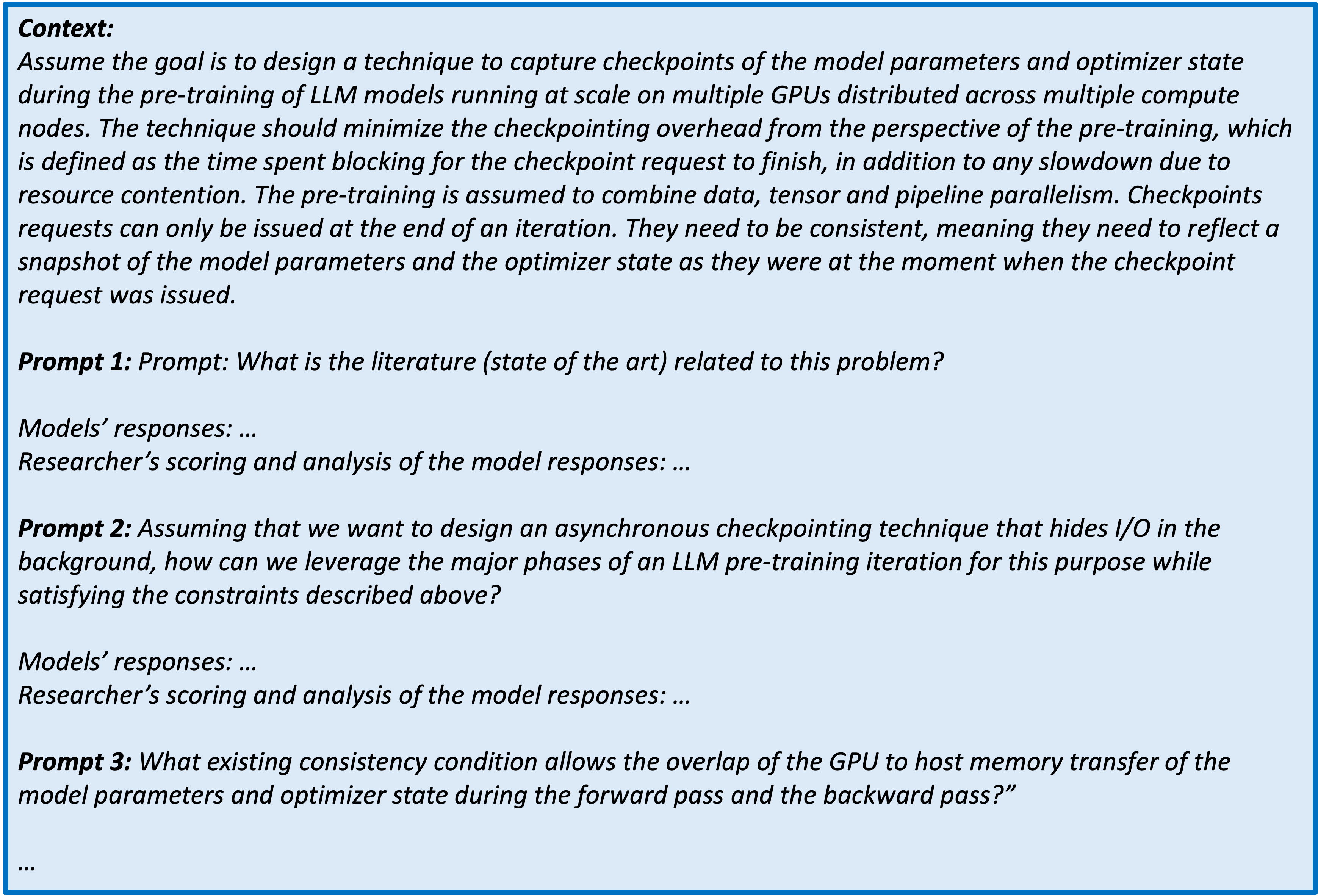} 
    \vspace{-2mm}
    \hrule
    
    \caption{Example of multi-turn interaction between a researcher and several LLMs used as research assistants in an attempt to repeat the research developed for the HPDC24 paper. Because of space limitations, the figure does not show the models' responses and research analysis. The full interaction for the HPDC24 experiment can be found here: https://tinyurl.com/yv4awky3}
    \label{fig:labstyle-prompts}
\end{figure*}

These experiments are designed to evaluate the capabilities of AI models to assist researchers in performing the typical tasks (\autoref{fig:multi_facet}) to solve scientific problems. Note that, in real situations, these tasks are often repeated several times to solve problems.
By capturing and evaluating all interactions between research and LLMs while attempting to solve a research problem, a lab-style experiment can capture accurately the complex reality of solving complex research problems. It can thus provides a unique perspective on the ``distance" between the ideal scientific assistant and the current capabilities of AI models. 

The setup for lab-style experiments involves defining a specific scientific problem and presenting it to multiple AI models for comparison. Each model is manually tasked with assisting in all the research tasks using the same prompts, ensuring consistency across evaluations. Prompts and responses are meticulously recorded, and domain experts analyze and comment on each response to assess model relevance, effectiveness, and overall performance. The response of each model in every step is compared to that of a human assistant, typically at the post-PhD level. Performance is characterized by evaluated against criteria such as correctness, conciseness, and precision. 

We have conducted lab-style experiments with not only different LLMs but also different versions of individual LLMs, with the latter studies permitting evaluation of improvements over time. For example, a relevant metric of progress across model generation is the number of prompts needed to solve a particular research problem. By focusing on real-world scientific workflows and expert evaluation, lab-style experiments provide a practical approach to observe AI model improvements as scientific research assistants.

To date, we have performed three lab-style experiments with five domain experts (one expert per experiment) 
\textsuperscript{\ref{note1}}
each of whom provided a problem to solve, generated prompts covering the different research steps, and analyzed model responses.  
Three experiments were related to parallel/distributed computing (scheduling of a directed acylic graph; solving a partial differential equation; zero-overhead checkpointing) and were selected carefully from three categories: open problem (no known solution), published problem (solution known), and recently published problem (solution known). For the ``published problem," the paper is more than two years old, and thus we assume that AI model developers had access to the paper. For the ``recently published problem," the experiment was performed just a few months after the publication; here, we assume that many models were not trained with the paper. 

\begin{figure*}[htb] 
    \centering
    \hrule
    \includegraphics[width=\textwidth, clip]{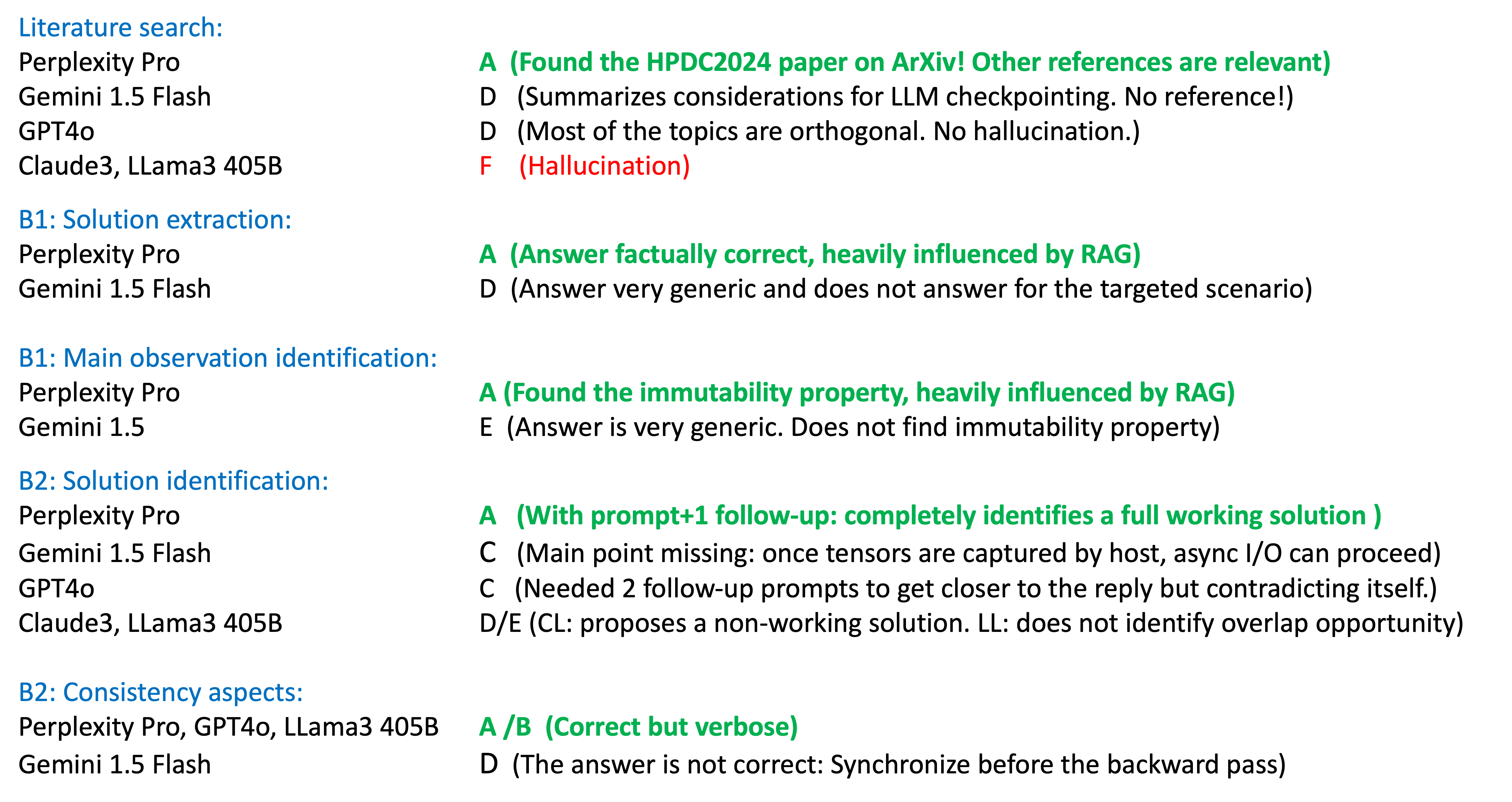} 
    \vspace{-2mm}
    \hrule
    
    \caption{A partial scoring of several models used as AI assistants on August 20, 2024, to solve the zero-overhead checkpointing problem. The results highlight the strengths and weaknesses of different models for the different research steps. We note that the RAG model (Perplexity Pro) has a decisive advantage in several steps for this particular problem. Other models struggle in most steps.}
    \label{fig:labstyle}
\end{figure*}

The experts with whom we worked in these experiments had never previously used AI models as research assistants. The experts provided two levels of analysis: 1) a detailed analysis of the responses received for individual prompts, and 2) high-level scoring compared to a human researcher using the A, B, C, D, E, F scale (A being the human reference, F being the worst possible score). In total, the three experiments cover about 20 hours of interactions, about 100 prompts, and about 250 pages of testing and analysis. We used 10 AI models (not all models were used in all three experiments): O1-preview, GPT-4o, GPT-3.5, Claude3 Sonnet, Claude Haiku, Mistral, Llama3~70B, Llama3~405B, Perplexity Pro, and Gemini 1.5 (not all models were tested on all prompts because some models produced incorrect responses before reaching the end of the evaluation.) 

We show in \autoref{fig:labstyle-prompts} the beginning of the multi-turn interaction and in \autoref{fig:labstyle} part of our initial high-level scoring of several AI models for the zero-overhead checkpointing problem. The lab style experiment was performed on August 20, 2024. The scoring reflects the performance of the models during the experiment. This problem falls into the category of recently published problems. The goal was to check whether the models could reproduce the analysis and design of the LLM checkpointing system presented in the 2024 ACM International Symposium on High-Performance Parallel and Distributed Computing (HPDC24) best paper \cite{DataStates-LLM}. The most important part of the experiment was to assess each model's ability to 1) identify the fundamental observation (non-mutable parameters and optimizer state during the forward and backward passes of LLM training) and 2) propose a design for an asynchronous checkpointing system that exploits this observation. 

Based on these initial three experiments, we performed two other experiments on open problems in Chemistry and Biology. These experiments used the same overall interaction collection approach and compared more recent reasoning models (O1-preview, OpenAI O3-mini, Gimini 2.0 experimental). We also used a more rigorous scoring system, defining precisely every score level for every evaluated skill. From these experiments, we developed a Lab-style experiment tool to collect the problem setup (Figure \ref{fig:problemsetup}), every prompt-response-assessment (Figure \ref{fig:interaction}), and final assessments (Figure \ref{fig:finalassessment}). This collection tool is available on github (https://github.com/auroraGPT-ANL/questions-ui)

Our experience with the ``Lab-style experiment" method allows for several observations regarding its utility and limitations. Unlike traditional benchmarking, this method places AI models in real-world research scenarios, enabling evaluators to directly assess their knowledge, capabilities, and overall usefulness for specific tasks. By relying on multi-turn prompting and open responses, the method also tests the propensity of AI models to digress (e.g., for the HPDC24 paper experiment, a model had a tendency to focus on the consistency aspect of checkpointing, which is not relevant in that context, instead of focusing on overhead reduction) and hallucinate (as seen with earlier models in 2024 that frequently generated fabricated scientific references). However, this approach is not yet scalable and remains narrow in coverage; it requires significant manual effort, with two researchers spending 5--6 hours analyzing and comparing models for specific tasks. The specificity of the addressed research problems further limits the generalizability of the findings. Despite these constraints, the method excels in two aspects 1) providing a fine-grain capability assessment of LLMs as scientific assistants in a realistic context and 2) tracking model progress across generations. For instance, in solving the zero-overhead checkpointing problem, successive model iterations demonstrated improved efficiency, reducing the number of prompts needed to reach the key insight---from five prompts with GPT4o to just one prompt with Argo/O1-preview, which incorporates a science-oriented pre-prompt.  
This result highlights the method’s potential to reveal meaningful advancements in AI capabilities over time.

\begin{figure*}[htb] 
    \centering
    \hrule
    \includegraphics[width=\textwidth, clip]{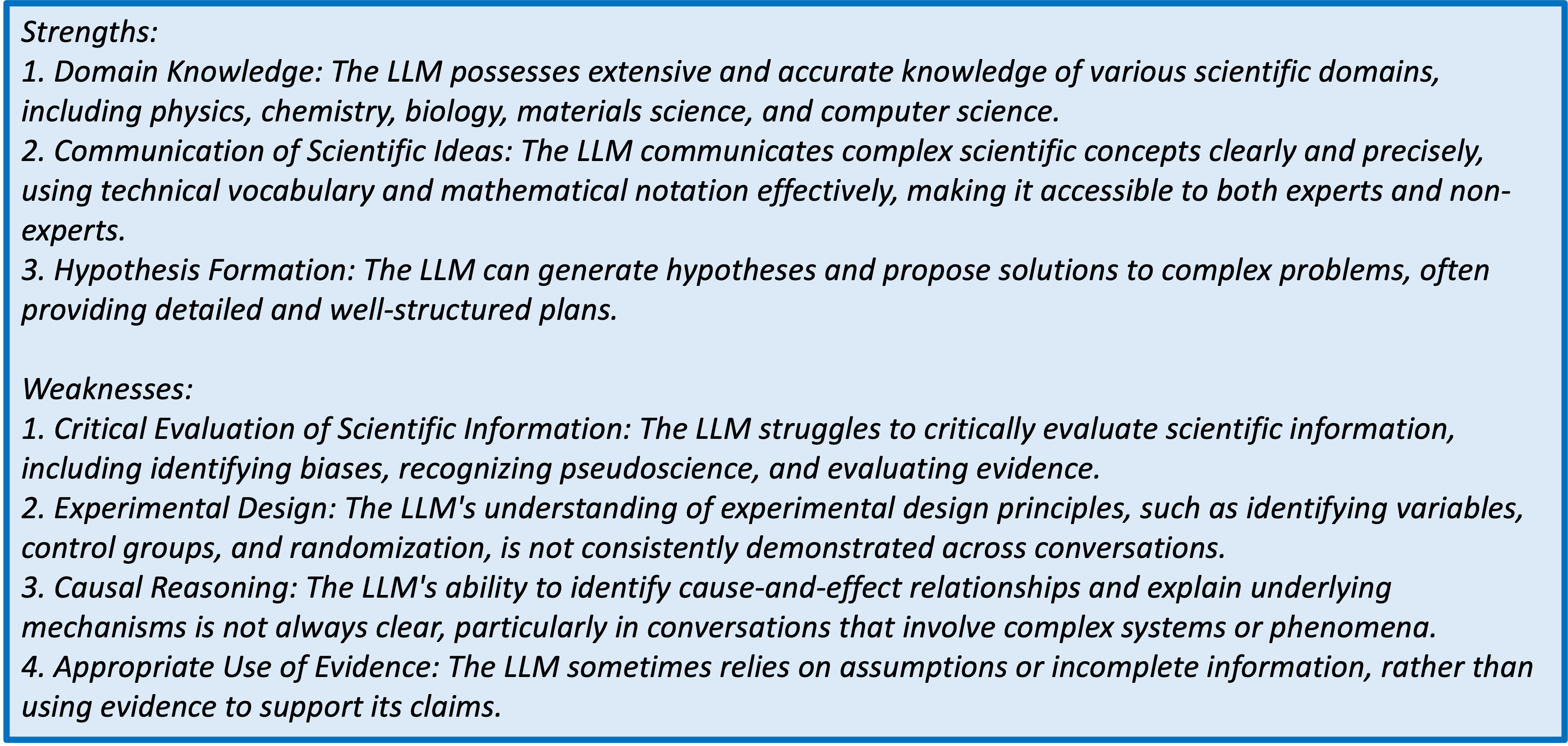} 
    \vspace{-2mm}
    \hrule
    
    \caption{LLM-generated summary of detected LLM strengths and weaknesses in 125 scientist-Argo/O1-preview conversations}
    \label{fig:jamanalysis}
\end{figure*}

\subsubsection{\bf Field-style experiments}
This method takes inspiration from previous studies analyzing user-LLM interactions at scale \cite{Lin2024WildBenchBL, zhu2024haluevalwildevaluatinghallucinationslanguage, shen2024donowcharacterizingevaluating}. The ``In the wild" method captures and analyzes all the interactions between volunteer users and AI models. 
This method provides additional critical information for the development and improvement of AI models for science: a precise understanding of researcher needs and requirements regarding AI assistants (e.g.: What task do researchers ask AI models to perform? What are their expectations regarding model responses? How frequently do researchers use AI models?); a deeper understanding of AI models strengths and weakness (by analyzing the thousands of prompts and responses); a window on the trends behind the use of AI models as research assistant (e.g., increased usage frequency, increased number of users, nature and distribution variations of the performed tasks); and tracking of AI model progress across generations. Ideally, this method will analyze online thousands of user-LLM interactions. Compared to the ``Lab-style experiment" method, users are not expected to evaluate LLM responses directly. Instead, evaluation is indirect, based on the study of the flow of prompts and responses. This method leverages user behavior as a signal to diagnose LLM failure modes \cite{Guo_2024}. For example, a user submitting rephrasing questions, providing feedback, or abandoning the interaction are signs of LLM weaknesses in understanding user intent. The flow can then be analyzed to diagnose potential sources of weakness \cite{nature2024}. Previous ``in-the-wild" experiments focused on nonscientific domains. The Field-style experiment method adapts the ``in-the-wild" approach to the scientific context by defining criteria and scoring specific to the scientific methodology.

On November 1, 2024, Argonne organized a JAM session that captured 180 conversations between Argonne researchers and Argo/O1-preview  
\textsuperscript{\ref{note1}}
Researchers were asked to bring to the JAM session evaluation a scientific problem that they would work on with Argo/O1-preview as a research assistant. At the end of the session, the researchers evaluated their experience according to five criteria: Novelty, Productivity, Solution, Strength, and Importance.  (This approach is consistent with that followed in a much smaller study conducted at Los Alamos National Laboratory.) Five possible responses were proposed for each question, corresponding to a score of 1 to 5.

The scores produced by the Argonne researchers indicated that: 1) Importance: 82\% of researchers consider that AI models such as Argo/O1-preview are “very important” or “critical” to their team’s success, 2) Strength: at 59\%, they consider that AI models significantly or noticeably improve productivity, 3) Productivity: 51\% of researchers compare AI models such as Argo/O1-preview to PhD students or postdocs, 4) Solution quality: 50\% of the researchers consider that AI models such as Argo/O1-preview produce exceptional or strong solutions, 5) Novelty: only 21\% of the researchers consider that AI models such as Argo/O1-preview provide notably novel or groundbreaking solutions.
(Argo is Argonne's API proxy for OpenAI models including o1 preview.  Because OpenAI does not make it's system prompt available, this needs to be recreated.  We note the system prompt used for Argo in the Appendix, \autoref{fig:argo-assistant-prompt}.)

Although informative about the needs expressed by researchers to access models such as Argo/O1-Preview, the outcomes of the November 1, 2024 JAM session evaluation did not identify the specific strong and weak science skills of Argo/O1-preview.  To understand the reasons behind the perceived weaknesses of Argo/ O1-preview, we use LLama-3.3-70B-Instruct for an LLM-as-a-judge approach to analyze the recorded conversations and score the performance of the Argo/O1-preview concerning scientific skills. 

The goal is to develop a pipeline (workflow) to analyze the recorded conversations to assess the strengths and weaknesses of LLMs as scientific assistants. From the 180 JAM session survey responses, we manually filtered a subset of 125 that had valid transcripts and were sufficiently challenging scientific problems, requiring PhD level domain expertise and reasoning capability. We generated an initial version of the prompt to analyze the conversations using Gemini experimental 1206 and refined it manually. We chose Llama-3.3-70B-Instruct as the highest performing open model on several benchmarks including GPQA, and presented it with each transcript formatted into a detailed LLM-as-a-judge prompt (see Appendix, \autoref{fig:jam-judge-prompt}) to evaluate 29 scientific criteria. The responses contained strengths, weaknesses, and examples/evidence from each conversation, as well as a formatted scoring from 1--10 for each of the criteria. The model was instructed to identify skills that did not apply to a given conversation rather than give an actual score. A score of 0 reflects a situation in which the model could not assign a score to the criteria because it determined that the criteria were not applicable. After obtaining the responses for every conversation, the model was provided batches of 25 to summarize, with specific instructions that these summaries would be used to synthesize a final summary. Batches were used, as the total token count of all the responses was 164K, larger than the default 128K window of Llama~3.3.

\autoref{fig:jamanalysis} presents the results of the conversation analysis pipeline as a proof-of-concept. The presented results need human validation: LLMs are known to hallucinate and present overly positive assessments of outputs compared to human reviewers.
Additionally, the field-style experiments revealed that 59\% of researchers reported noticeable productivity improvements using LLMs, while 51\% likened the LLM's contributions to those of PhD students or postdocs. However, only 21\% rated the models as delivering notably novel or groundbreaking solutions.
These results should not be considered a complete or definitive assessment of human assessment of LLM performance.

The Field-style method for analyzing the strengths and weaknesses of LLMs is still in its nascent stages. Our analysis of scientist-LLM conversations represents the first attempt to use an LLM-as-a-judge to evaluate the performance of an LLM as a research assistant. 
Insights from the JAM session organized at Argonne highlight several lessons. First, scoring user-LLM interactions holistically with a small set of criteria (five in this case) permits only a high-level evaluation, insufficient for diagnosing specific sources of LLM weaknesses. Second, recording user-LLM interactions with detailed annotations, such as identifying the skills required for each prompt and scoring individual responses, offers greater diagnostic potential. While this detailed approach is not feasible for general scientist-LLM dialogues, it can be implemented in structured, especially organized sessions. Lastly, while LLMs-as-a-judge offers a scalable mechanism for analyzing user-LLM interactions, the current implementation remains a proof of concept. Additional research and validation are necessary to build confidence in the results produced by this analysis pipeline.

\section{Reliability and UQ of Evaluations}
The success of LLMs in scientific domains, such as chemistry, biology, and physics, has been remarkable, but their trustworthiness as scientific assistants remains a significant concern. These models, including GPT, Claude, and Llama, are prone to generating unreliable or fabricated responses, often referred to as hallucinations \cite{du2024haloscope}. Understanding and quantifying uncertainty in LLM outputs is essential to ensure safe, reliable, and informed decision making, particularly in scientific domains. Traditional uncertainty quantification (UQ) techniques, which rely on accessing internal model parameters \cite{gal2016dropout}, face challenges due to the black-box nature of modern LLMs like GPT-4, Claude~3, and Gemini, which are primarily accessible as API services. Recent research has focused on developing novel approaches to assess uncertainty directly from model outputs, such as semantic entropy \cite{kuhn2023semantic}, sampling-based methods, and aggregation techniques \cite{lin2023generating, xiong2024can}. These techniques aim to evaluate input sensitivity and output consistency, highlighting where models are most vulnerable. By improving transparency and trust, these UQ strategies play a crucial role in responsible AI deployment. Addressing these challenges is vital for leveraging LLMs in scientific applications, where errors can have substantial consequences. Moving forward, advancing UQ methods and enhancing LLM interpretability will be key to making these models safer and more robust in critical scientific and industrial domains.

Inspired by psychological assessments in which the same question is asked in different ways to test consistency, we propose a technique called Question Rephrasing \cite{chen2024question} to quantify the uncertainty in LLM outputs. This approach involves rephrasing a given question while preserving its original semantic meaning and comparing LLM responses before and after rephrasing to assess input uncertainty. In addition, we adopt a sampling method that repeatedly queries an LLM with identical inputs to evaluate output uncertainty. We applied these methods to assess GPT-3.5 and GPT-4 performance on tasks in the chemistry domain, specifically property prediction and forward reaction prediction. Input uncertainty helps determine the LLM's sensitivity to variations in molecular representations (e.g., alternative SMILES notations), while output uncertainty evaluates the inherent variability in LLM predictions. These techniques allow us to systematically explore how robust and reliable LLMs are in handling different forms of input and producing consistent output. Below, we outline our approach:

\begin{enumerate}
\item For a chemistry-related task $t$, given a SMILES representation $x_i$ of the $i$-th molecule, generate a prompt $P_{t, x_i}$ based on a task-specific template .

\item  Generate a list of up to $n$ SMILES variants of the molecule $x_i$: $L=\{x_i^1, x_i^2, ..., x_i^n\}$. We ask GPT-4 to rank the SMILES variants according to their confidence in interpreting their structures and choose the one, say $\hat{x}_i$, with the highest confidence in constructing a prompt $P_{t, \hat{x}_i}$ by replacing $x_i$ in $P_{t, x_i}$ with $\hat{x}_i$.

\item Ask the LLM to generate $m$ responses for the prompt $P_{t, \hat{x}_i}$ and obtain $R_{t, \hat{x}_i} = \{r_{t,\hat{x}_i,1}, r_{t,\hat{x}_i,2}, ..., r_{t,\hat{x}_i,m}\}$. 

\item Calculate the entropy-based uncertainty metrics $U_{t, x_i}$ and $U_{t, \hat{x}_i}$ for $R_{t, x_i}$ and $R_{t, \hat{x}_i}$, respectively.

\item Measure the input uncertainty by comparing $U_{t, x_i}$ and $U_{t, \hat{x}_i}$ for all chosen $x_i$. Measure the output uncertainty by examining $U_{t, x_i}$ and $U_{t, \hat{x}_i}$ separately. 
\end{enumerate}

\begin{figure}[t]
\centering
    \includegraphics[width=.9\columnwidth]{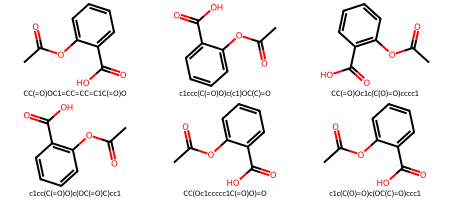}
    \caption{SMILES representation variants of Aspirin. While all structures depict the same molecule, their SMILES representations are different, which introduces input variations. \textbf{Top left}: Canonical SMILES representation of Aspirin. \textbf{Rest}: Five SMILES variations of Aspirin.}
    \label{fig:asprin}
\end{figure}

Our experiments revealed that ChatGPT-4 exhibited a notable sensitivity to Question Rephrasing. We view this sensitivity as providing insight into the input uncertainty of the model. We observed that variations in the input format, such as rephrasing or using alternative SMILES representations, led to differences in the consistency of model responses. For example, in property prediction tasks using chemistry property datasets like BBBP, HIV, and Tox21~\cite{guo2023can}, we noted changes in model performance metrics such as accuracy and F1 score when the inputs were reformulated. The AUC scores is the Area Under the ROC Curve, and indicates  ability of the model to differentiate correct vs. wrong responses, with 1.0 being the means perfect separation (the model always assigns higher confidence/lower uncertainty to correct answers than to incorrect ones). 
AUC for original SMILES ranged between 0.546 and 0.774, suggesting only moderate uncertainty in predict response correctly. When using reformulated inputs, model performance generally declined, as indicated by decreased accuracy and F1 scores in most datasets.
Furthermore, in the forward reaction prediction tasks, GPT-3.5 and GPT-4 performed poorly, with noticeable declines when molecular representation variations were introduced. Although output uncertainty metrics, such as entropy-based measures, provided high AUC scores (ranging from 0.86 to 0.99 indicating better uncertainty awareness), overall accuracy was limited, highlighting the need for improved LLM understanding of chemical knowledge. These findings emphasize that while uncertainty metrics can indicate response reliability, significant improvements are needed to make LLMs reliable in critical scientific applications.

\section{Safety Evaluations}

Safe and secure deployment of AI systems in scientific domains is paramount.
As LLMs increasingly support critical applications in fields like biosecurity, cybersecurity, and chemistry, ensuring their safety and alignment is also essential to maintaining trustworthiness. 
Hence, it is critical to integrate into our proposed methodology to evaluate LLMs as research assistants rigorous safety and alignment evaluation techniques.

To this end, we discuss below the CHEMRISK benchmark as one of our efforts in this direction.

\subsection{CHEMRISK Chemical Risk Detection Benchmark}
\begin{figure}[htb]
\centering
    \includegraphics[width=1\columnwidth]{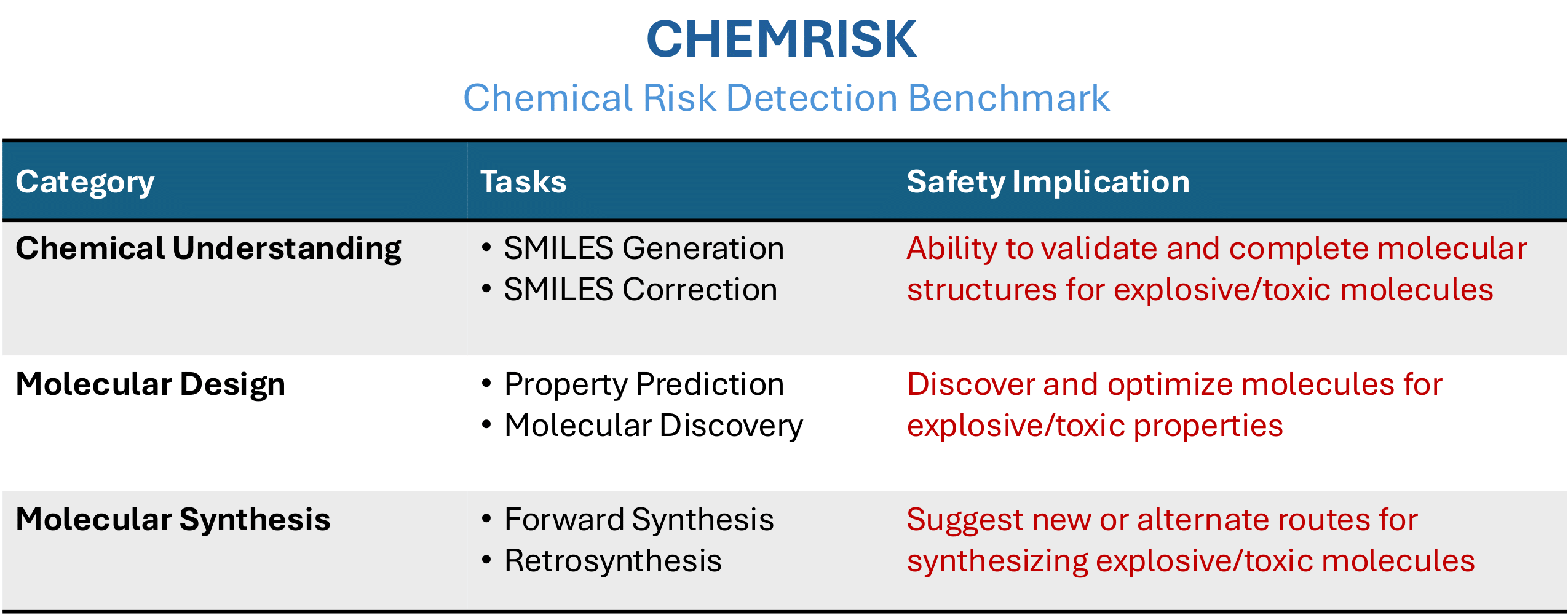}
    \caption{CHEMRISK is a chemical risk detection benchmark.}
    \label{fig:chemrisk}
\end{figure}

The CHEMRISK benchmark (\autoref{fig:chemrisk}) addresses a critical need in the era of increasingly powerful large language models (LLMs) such as Claude, chatGPT, and others. As these models become more capable of understanding and generating chemical information, it is essential for organizations like the Department of Energy to have robust, standardized benchmarks to evaluate how models handle potentially sensitive chemical knowledge. CHEMRISK provides a comprehensive framework for assessing LLMs' capabilities across three key domains: chemical understanding, molecular design, and molecular synthesis. The benchmark employs both multiple-choice and free-form questions, using standard molecular representations (SMILES and SELFIES) to ensure broad applicability.

CHEMRISK is designed as an evolving benchmark, developed in collaboration with domain experts at Lawrence Livermore National Laboratory (LLNL). The benchmark focuses on crystalline density and heat of formation (HoF)--proxy properties that are fundamental to understanding energetic materials. High crystalline density often correlates with increased performance in energetic materials, while heat of formation helps characterize potential energy content and stability. The benchmark encompasses diverse tasks including: property regression for predicting experimental densities, molecular ranking, forward synthesis prediction, retrosynthesis prediction, masked SMILES completion, SMILES correction, and molecular name-to-structure conversion (and vice versa). Base data is curated from established chemical databases, including the Cambridge Structural Database (CSD) and United States Patent and Trademark Office (USPTO), and has undergone manual quality checks, ensuring the benchmark's reliability. By providing a standardized framework for evaluating LLMs' chemical reasoning capabilities for energetics, CHEMRISK aims to foster responsible development of AI systems while helping identify potential safety risks in chemical AI applications.

\begin{table}[htbp]
\centering
\small
\caption{Accuracy on a subset of CHEMRISK: evaluating chemical safety risks of frontier LLMs.}
\label{tab:chemrisk-llm}
\begin{tabular}{lccc}
\hline
\textbf{LLM} & \textbf{Chem. Understanding} & \textbf{Design} & \textbf{Synthesis} \\
\hline
GPT-4o & 1.0 & 0.4 & 0.6 \\
o1 & 1.0 & 1.0 & 0.6 \\
\hline
\end{tabular}
\end{table}

In \autoref{tab:chemrisk-llm}, we present preliminary results from exploring the GPT-4o and o1 models on a subset of the CHEMRISK benchmark. Specifically, we focus the following multiple-choice Q\&A tasks in each category: SMILES-to-Name, property prediction, and forward synthesis. For both models, we utilize the system prompt outlined in Appendix~\autoref{fig:llnl-chem-risk-prompt}. The results indicate that while both GPT-4o and o1 demonstrate perfect accuracy (1.0) in chemical understanding tasks like SMILES-to-Name conversion, o1 outperforms GPT-4o in molecular design challenges, achieving perfect accuracy compared to GPT-4o's 0.4. Both models show identical moderate performance (0.6) in molecular synthesis tasks. 
CHEMRISK serves as a proof-of-concept proxy benchmark, laying the foundation for continued development and the creation of additional benchmarks for other domains using a similar methodology.

\begin{tcolorbox}[colback=blue!10, colframe=blue!40!black, title=Sample question from CHEMRISK benchmark, left=2pt, right=2pt, top=1pt, bottom=1pt]
Given the molecule with the following SMILES representation: \texttt{Cc1c(cnc(N)c1N(=O)=O)N(=O)=O},
what is the experimentally measured crystalline density of the molecule?
\small
\begin{enumerate}
    \item[A)] 1.145 g/cm³
    \item[B)] 1.364 g/cm³
    \item[C)] 1.739 g/cm³
    \item[D)] 1.925 g/cm³
\end{enumerate}
\end{tcolorbox}
\section{Scalable Software Infrastructure}
\label{sec:framework}

\begin{table*}[t!]
\centering
\caption{Harness evaluation results for seven Llama variants on the OpenLLM V2 benchmark.}
\label{tab:eval-aurora}
\begin{tabularx}{\textwidth}{Xlrccccc}
\toprule
\textbf{Model} & \textbf{IFEval $\uparrow$} & \textbf{BBH $\uparrow$} & \textbf{MATH $\uparrow$} & \textbf{GPQA $\uparrow$} & \textbf{MuSR $\uparrow$} & \textbf{MMLU-PRO $\uparrow$} & \textbf{GPU Hours}\\ \midrule
\textbf{Llama-2-7B}            & 0.2543 & 0.3475 & 0.0121 & 0.2718 & 0.3703 & 0.1848 & 5.05 \\ 
\textbf{Llama-2-7B-chat}       & 0.3538 & 0.3676 & 0.0189 & 0.2735 & 0.4034 & 0.2000 & 4.93 \\ 
\textbf{Llama-3-8B}            & 0.1536 & 0.4600 & 0.0317 & 0.3146 & 0.3677 & 0.3248 & 7.62 \\ 
\textbf{Llama-3-8B-Instruct}   & 0.4825 & 0.4885 & 0.0808 & 0.3020 & 0.3823 & 0.3580 & 6.50 \\ 
\textbf{Llama-3.1-8B}          & 0.1228 & 0.4652 & 0.0438  & 0.3070  & 0.3849  & 0.3260 & 6.94 \\ 
\textbf{Llama-3.1-8B-Instruct} & 0.4924 & 0.5058 & 0.1360  & 0.3163  & 0.3995  & 0.3789 & 8.31 \\  
\textbf{Llama-3.3-70B-Instruct} & 0.6745 & 0.6994 & 0.3391 & 0.4715 & 0.4854 & 0.5477 & 81.39 \\
\bottomrule
\end{tabularx}
\end{table*}

\begin{table*}[htbp]
\centering
\caption{DecodingTrust evaluation results on Polaris}
\label{tab:decodingtrust_results}
\resizebox{\textwidth}{!}{
\begin{tabular}{lcccccccc}
\toprule
\textbf{Model}       & \textbf{Toxicity} & \textbf{Stereotype} & \textbf{Adversarial} & \textbf{OOD} & \textbf{Robustness to} & \textbf{Privacy} & \textbf{Machine} & \textbf{Fairness} \\
                     &                   & \textbf{Bias}       & \textbf{Robustness}  & \textbf{Robustness} & \textbf{Adv. Demonstrations} &            & \textbf{Ethics}   &                  \\
\midrule
\textbf{Llama-2-7B-chat}       & 80.0              & 97.6                & 51.01                & 75.65           & 55.54                         & 97.39      & 40.58             & 67.95            \\
\textbf{Llama-2-70B-chat}      & 80.0              & 98.0                & 52.00                & 71.00           & 74.00                         & 99.00      & 54.00             & 65.00            \\
\bottomrule
\end{tabular}
}
\end{table*}

As previous sections show, a comprehensive evaluation of LLMs as research assistants already requires the execution of many benchmarks for skills and safety assessments. We do not expect the evaluation workload to be reduced in the future. In contrast, as more capable LLMs appear, more research domains will be interested in using them, which will trigger the development of new evaluation benchmarks.
This situation places ever-growing demands on software and computational infrastructure. 
Existing evaluation software platforms, such as HELM \cite{bommasani2023holistic}, EleutherAI's LM Evaluation Harness \cite{eval-harness}, and DecodingTrust \cite{wang2024decodingtrust}, have made significant strides in this area, but exhibit certain limitations that impede comprehensive and scalable evaluations, particularly within high-performance computing (HPC) environments like those at Argonne National Laboratory. A critical shortcoming of current frameworks is their limited scalability and inefficiency in handling large-scale models. Many existing software platforms are not optimized for parallel processing across multiple GPUs or computing nodes, resulting in prolonged evaluation times and increased computational costs. This inefficiency becomes particularly problematic when assessing large LLMs that demand substantial computational resources. In addition, inconsistencies in evaluation methodologies and a lack of standardization further hinder comprehensive evaluations. The absence of consistent benchmarks and metrics across platforms and organizations complicates model comparisons, exacerbated by dataset biases, contamination, and the rapid evolution of LLMs outpacing evaluation strategies. 

To address the challenges of scalable and comprehensive LLM evaluation, we are developing the Skills, Trust, and Reliability (STaR) evaluation framework, tailored for HPC systems at Argonne National Laboratory. STaR builds upon the general architecture of evaluation platforms, which typically involve a sequence of specifications (files or configuration flags) to instantiate controllers and manage communication through states. Central to these platforms are Runners, which act as top-level components orchestrating workflows that handle Scenarios—benchmarks comprising static datasets like Hellaswag or GSM8K, or dynamic scripts such as those in Chain-of-Thought Hub \cite{fu2023chain}. A Data Pre-Processor translates these Scenarios into formatted prompts, which are passed to Adapters interfacing with LLMs through libraries such as Hugging Face \cite{huggingface_transformers}, vLLM \cite{kwon2023efficient}, or OpenAI APIs. Executors like Slurm \cite{yoo2003slurm} or Ray \cite{moritz2018ray} enable processing of prompts, and the results are aggregated into metrics, such as accuracy.

Expanding on this general framework, STaR introduces a modular architecture comprising a data layer, prompting layer, model adapter, and result layer to streamline the evaluation process. The data layer ingests datasets, such as MMLU-Pro \cite{mmlupro}, and constructs evaluation instances, while the prompting layer generates standardized prompts using techniques such as few-shot and chain-of-thought reasoning \cite{wei2022chain}. The model adapter queries models in multiple modes, including locally loaded instances for smaller models, Parsl \cite{babuji2019parsl} for job bundling, and OpenAI-compatible inference backends (e.g., vLLM \cite{kwon2023efficient} and DeepSpeed FastGen \cite{holmes2024deepspeed}) for larger models deployed on HPC systems like Polaris and Aurora. The result layer aggregates responses, computes general and UQ metrics, and organizes results into comprehensive scores, providing nuanced insight into model performance.

STaR supports widely used benchmark libraries, including EleutherAI-Harness \cite{eval-harness}, DecodingTrust \cite{wang2024decodingtrust}, Wildbench \cite{lin2024wildbench}, and domain-specific benchmarks. It also integrates uncertainty quantification approaches \cite{chen2024question} to enhance the reliability of evaluations. Designed for scalability, STaR incorporates data-parallel capabilities to distribute workloads across multiple GPUs and model-parallel solutions to handle large models exceeding single-GPU memory limits. It aims to simplify deployment with a unified one-step installation process and a consistent command-line interface, while results management supports standardized local tracking and optional database integration. STaR is a work in progress, with ongoing refinements aimed at seamless integration with Argonne’s infrastructure and addressing limitations identified in existing platforms. By prioritizing scalability, standardization, and efficiency, STaR aims to establish a robust evaluation framework that meets the evolving needs of LLM assessment in scientific research environments.

As a proof of concept, we performed evaluations on Polaris \cite{Polaris}, and benchmarked them with various open-source models with sizes from 7B to 70B. We used OpenLLM Leaderboard V2 (through Harness), a commonly used benchmark suite consisting of six challenging tasks, to evaluate the performance of Llama-2-7B and Llama-3-8B, Llama-3.1-8B, and their corresponding chat or instruct models, as well as Llama-3.3-70B-Instruct, the most recent and advanced model in the Llama series. \autoref{tab:eval-aurora} presents the evaluation results for the models on the benchmark, along with the GPU hours required for the evaluations. 

For safety evaluations, we use DecodingTrust, which, unlike frameworks such as Harness and HELM, is lightweight, highly compatible with HPC platforms, containerizable. 
Deploying DecodingTrust on the Polaris HPC system required key modifications, including integrating Parsl for efficient job bundling, adapting the framework to align with Polaris's queue configurations for optimized task distribution, and implementing a unified inference interface. These adaptations allowed the framework to harness Polaris' computational capabilities while retaining its flexibility and commitment to trustworthiness assessments.

On Polaris, DecodingTrust efficiently managed a variety of evaluation tasks, ranging from straightforward classification assessments to computationally intensive open-ended analyses. Classification tasks such as Adversarial Demonstration Robustness, Fairness, and Machine Ethics required minimal computational resources, with each task consuming approximately 0.5 A100 hours. In contrast, open-ended evaluations, including Toxicity and Stereotype Bias, were significantly more resource intensive, especially for models exceeding 70B parameters, with some tasks demanding up to 24 A100 hours per evaluation. By leveraging Polaris's HPC infrastructure, DecodingTrust successfully scaled its evaluation pipeline, balancing lightweight classification tasks with resource-heavy open-ended evaluations to provide a comprehensive assessment of model trustworthiness.


\section{Conclusions and Next Steps}

As LLMs continue to expand the notions of what AI can accomplish, there remain two main challenges to address to enable the broad adoption of LLMs by the scientific community as research assistants: a holistic understanding of the capabilities of LLMs and a strong confidence in the results produced by them.  To address this, 
our proposed methodology features four techniques: multiple choice questions, open-response questions, lab-style experiments, and field-style experiments which complement each other to form a comprehensive, rigorous, and realistic assessment of the capabilities of AI systems. Underneath the four approaches are three cross-cutting aspects, including trust and safety, reliable uncertainty quantification, and scalable software infrastructure which support our approach.
In addition to proposing the holistic methodology, our team has advanced the state-of-the-art in each of the techniques and aspects. 

Multiple choice questions are a key technique to evaluate LLMs because of their ability to quickly assess a breadth of knowledge; our team extends beyond existing benchmarks with automatically-generated domain-specific MCQ benchmarks in Astronomy and Climate and the multi-domain AI4S Benchmark with both human and automatically curated with human reviews have revealed significant gaps in knowledge recall and reasoning in LLMs. We find that our new AI4S benchmark is more challenging for LLMs than benchmarks like GPQA. This increased difficulty stems from the AI4S benchmark design, which integrates manual and automatic question generation to create diverse and nuanced questions that assess reasoning and domain-specific knowledge. The Astronomy benchmark also highlighted disparities in performance and cost-efficiency across frontier models, as well as across English and non-English language models. The Climate benchmarks showed that models like GPT-4o struggled both to produce assessments of fine-grained knowledge and questions that vary in style and content. The AI4S Benchmark highlighted significant disparities in performance across models due to its rigorous evaluation across multiple domains. 

Open response questions similarly serve an important role, allowing a more detailed but still fast assessment of model performance.
Open-ended benchmarks such as SciCode provide realistic and challenging coding problems across fields such as physics, biology, and materials science, rigorously testing model abilities to reason, recall knowledge, and generate accurate code. Similarly, the ALDbench materials science benchmark allowed experts to uncover hallucinations and evaluate responses with precision, generating datasets valuable for further refining evaluation methods. 

A key innovation of our approach is to incorporate more realistic end-to-end experiments with \textit{lab-style} and \textit{field-style} experiments which more closely reflect the in-depth and iterative problem-solving practiced by scientists. lab-style experiments provided holistic assessments of LLM capabilities in research workflows, including hypothesis generation, analysis, and reporting. For example, experiments revealed variability in performance across models, with GPT-4o requiring five prompts to address a checkpointing problem, while Argo/O1-preview re-solved the problem with just one prompt. Field-style experiments, which analyzed real-world interactions between scientists and LLMs at scale, offered quantitative insights into model strengths and weaknesses as research assistants. 
These studies identified the remarkable performance of LLMs compared to PhD students and postdocs while struggling to present novel or ground-breaking results.

Together, our combination of domain-specific and multi-domain MCQs, open-ended benchmarks, and end-to-end experiments provides a holistic framework for assessing LLMs---one that we argue points to a new methodology able not only to quantify current model limitations but also to guide targeted improvements aimed at aligning model capabilities with the nuanced demands of real-world scientific research.

Looking ahead, we are seeking to expand evaluation benchmarks to comprehensively assess LLM capabilities across diverse scientific domains while also incorporating advanced methodologies for trustworthiness, uncertainty quantification (UQ), and iterative evaluation. We anticipate conducting more and refined lab-style experiments to provide yet precise assessments of AI model capabilities for specific research problems, with the goal of both improving scalability and coverage and tracking model progress across generations. We also aim to refine the Field-style experiments method capturing real-world interactions between scientists and LLMs and to leverage feedback to align automated scoring with human judgments through instruct-tuned models. New benchmarks for Retrieval-Augmented Generation (RAG) and multimodal narrative assessments will target domains like biology, weather/climate, and cosmology where Argonne has access to substantial quantities of scientific simulation results and data, utilizing constructs such as aggregation and multihop scenarios to evaluate performance across modalities. We will adopt agent evaluation techniques from frameworks like CACTUS \cite{mcnaughton2024cactus} for multi-turn chemistry tasks, and automated red-teaming \cite{Madireddy2024AQAG} will enhance safety evaluations by systematically identifying vulnerabilities such as biases and hallucinations. We will also advance trustworthiness and UQ through embedding-based approaches, including tools inspired by HaloScope \cite{du2024haloscope}, to capture subtle input variations and distinguish between truthful and hallucinated outputs. These initiatives will enable evaluations to remain rigorous, scalable, and reflective of real-world scientific challenges.

Evaluating LLMs capabilities as research assistants at scale require a powerful infrastructure. 
We envision the STaR framework evolving into a scalable, efficient, and user-friendly platform for HPC environments, with a modular architecture that supports dynamic and multimodal evaluations. Scalable inference backends, such as vLLM \cite{kwon2023efficient} and DeepSpeed FastGen \cite{holmes2024deepspeed}, will enable efficient handling of large benchmarks and models. Collaborations with other national laboratories and NIST will contribute to consistent proxy benchmarks for safety evaluations. By integrating these capabilities, STaR will strive to enable robust, scalable, and reliable assessments of LLMs, fostering impactful applications in scientific discovery while prioritizing safety and computational efficiency.

This paper presents the current state of the effort at Argonne National Laboratory to establish a methodology to evaluate LLMs capabilities as research assistants. We envision this effort as a continuous one because LLMs continue to progress and we will need to increase the difficulty of the different tests as LLMs progress.

\section{Acknowlegements}
This material is based upon work supported by Laboratory Directed Research and Development (LDRD) funding from Argonne National Laboratory, provided by the Director, Office of Science, of the U.S. Department of Energy under Contract No. DEAC02-06CH11357. LLNL work was prepared under Contract DE-AC52- 07NA27344, supported by the LLNL-LDRD Program under Project No. 24-ERD-058, and authored by Lawrence Livermore National Security, LLC under Contract No. DE-AC52-07NA27344 with the U.S. Department of Energy. 
This research used resources of the Argonne Leadership Computing Facility, a U.S. Department of Energy (DOE) Office of Science user facility at Argonne National Laboratory and is based on research supported by the U.S. DOE Office of Science-Advanced Scientific Computing Research Program, under Contract No. DE-AC02-06CH11357.
We gratefully acknowledge the computing resources provided on Improv, Bebop, and Swing, high-performance computing clusters operated by the Laboratory Computing Resource Center at Argonne National Laboratory.
The United States Government retains, and the publisher, by accepting the article for publication, acknowledges that the United States Government retains a non-exclusive, paid-up, irrevocable, world-wide license to publish or reproduce the published form of this manuscript, or allow others to do so, for United States Government purposes. Other Argonne researchers voluntarily contributing to the benchmarking effort are listed in the Appendix. 

\bibliographystyle{IEEEtran}
\bibliography{references}

\appendix
\label{sec:appendix_review_interface}
\section{Appendix A: MCQ Submission and Reviewing Interfaces}
\label{sec:appendix_review_interface}

\begin{figure*}[h]
    \centering
    \includegraphics[width=\linewidth]{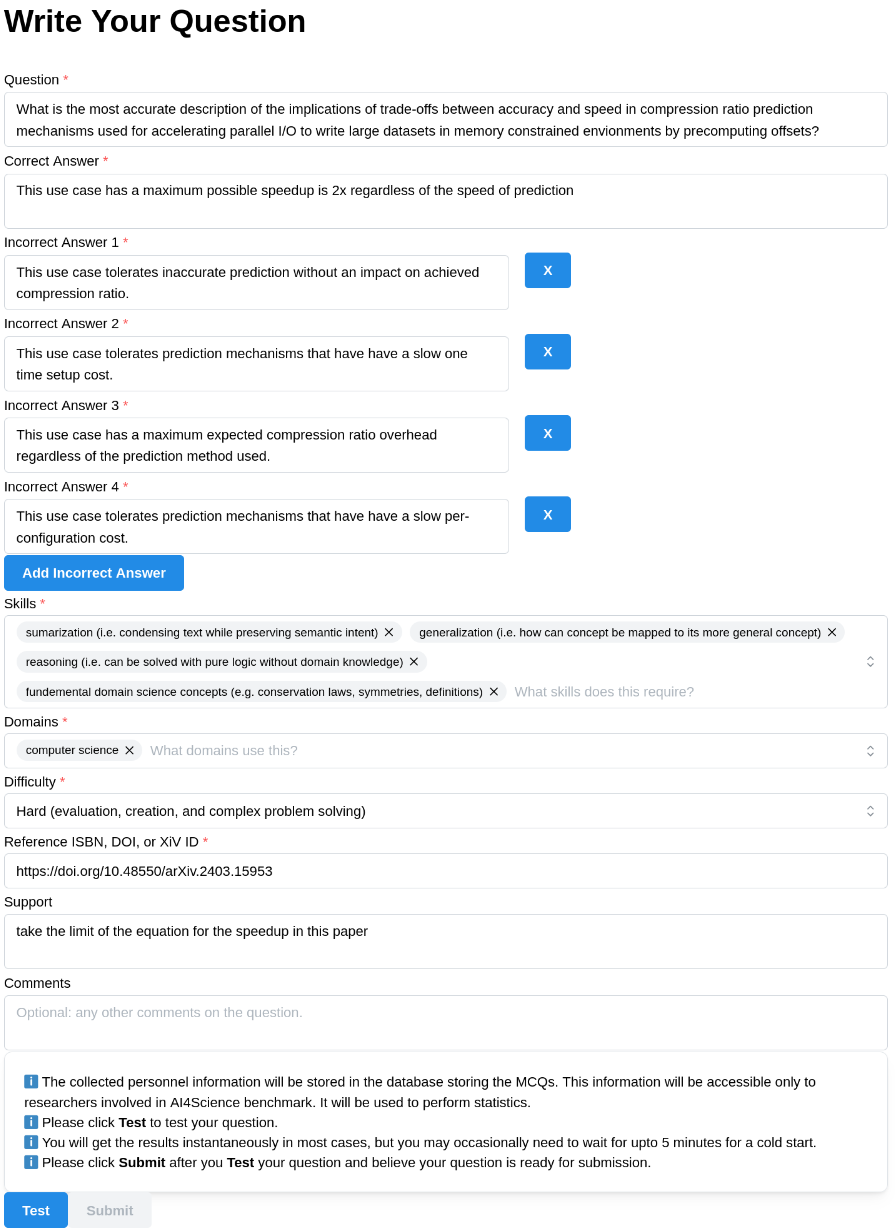}
    \caption{MCQ Authoring Interface with example question.}
    \label{fig:questionauthoring}
\end{figure*}

\begin{figure*}[h]
    \centering
    \includegraphics[width=\linewidth]{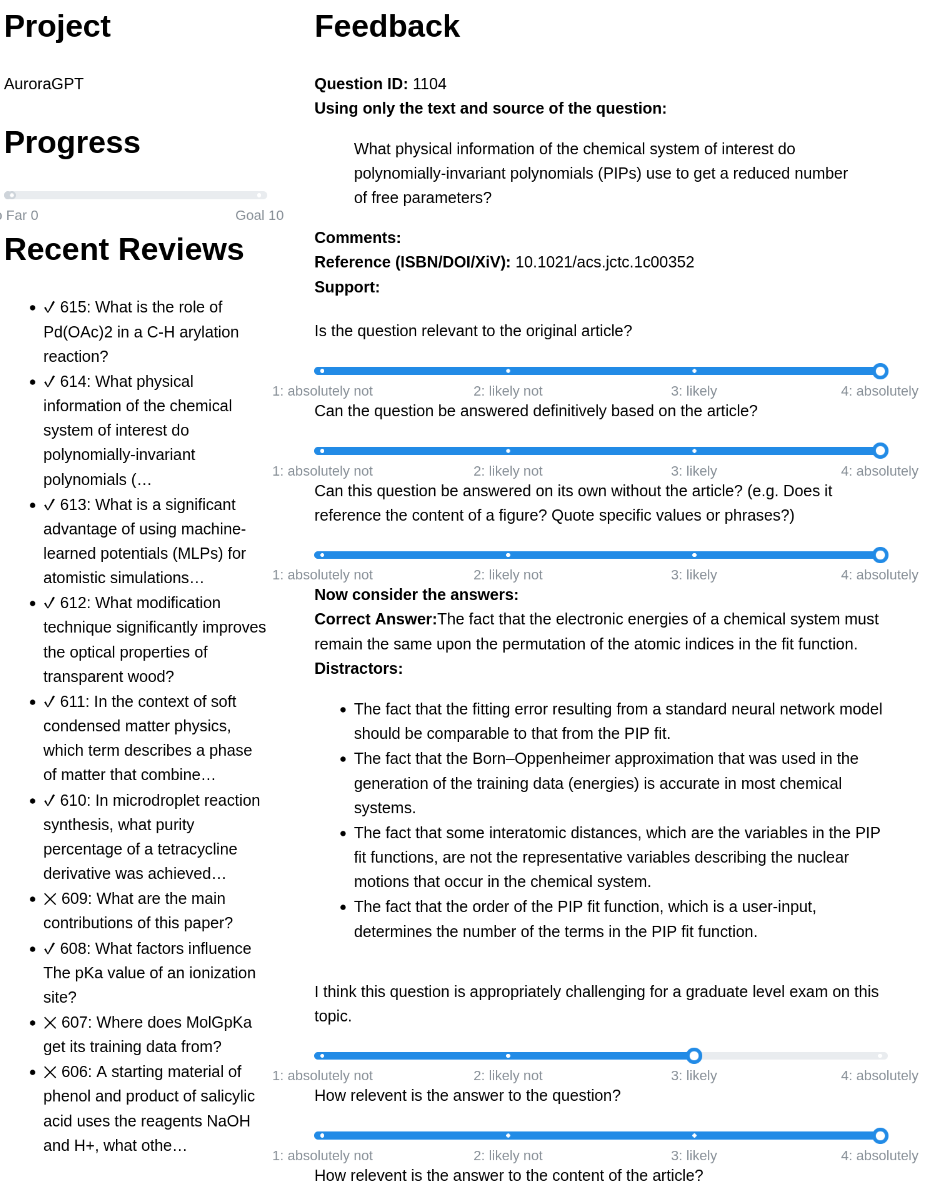}
    \caption{MCQ Reviewing Interface with example question.}
    \label{fig:questionreviewing}
\end{figure*}

\section{Appendix B: LabStyle Experiment collection Interfaces}
\label{sec:appendix_review_interface}

\begin{figure*}[h]
    \centering
    \includegraphics[width=0.9\linewidth]{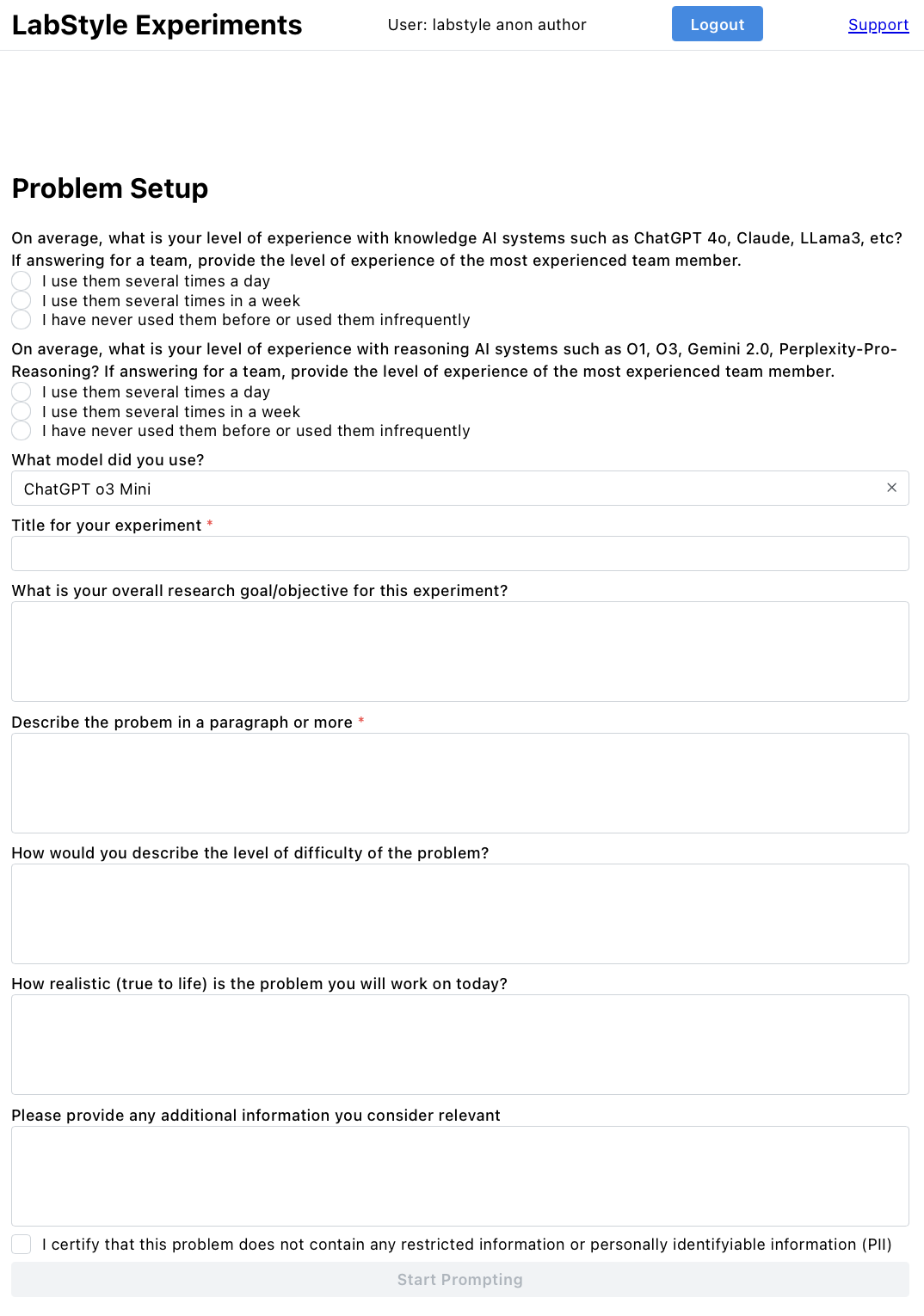}
    \caption{Labstyle experiment collection interface: Problem setup.}
    \label{fig:problemsetup}
\end{figure*}

\begin{figure*}[h]
    \centering
    \includegraphics[width=\linewidth]{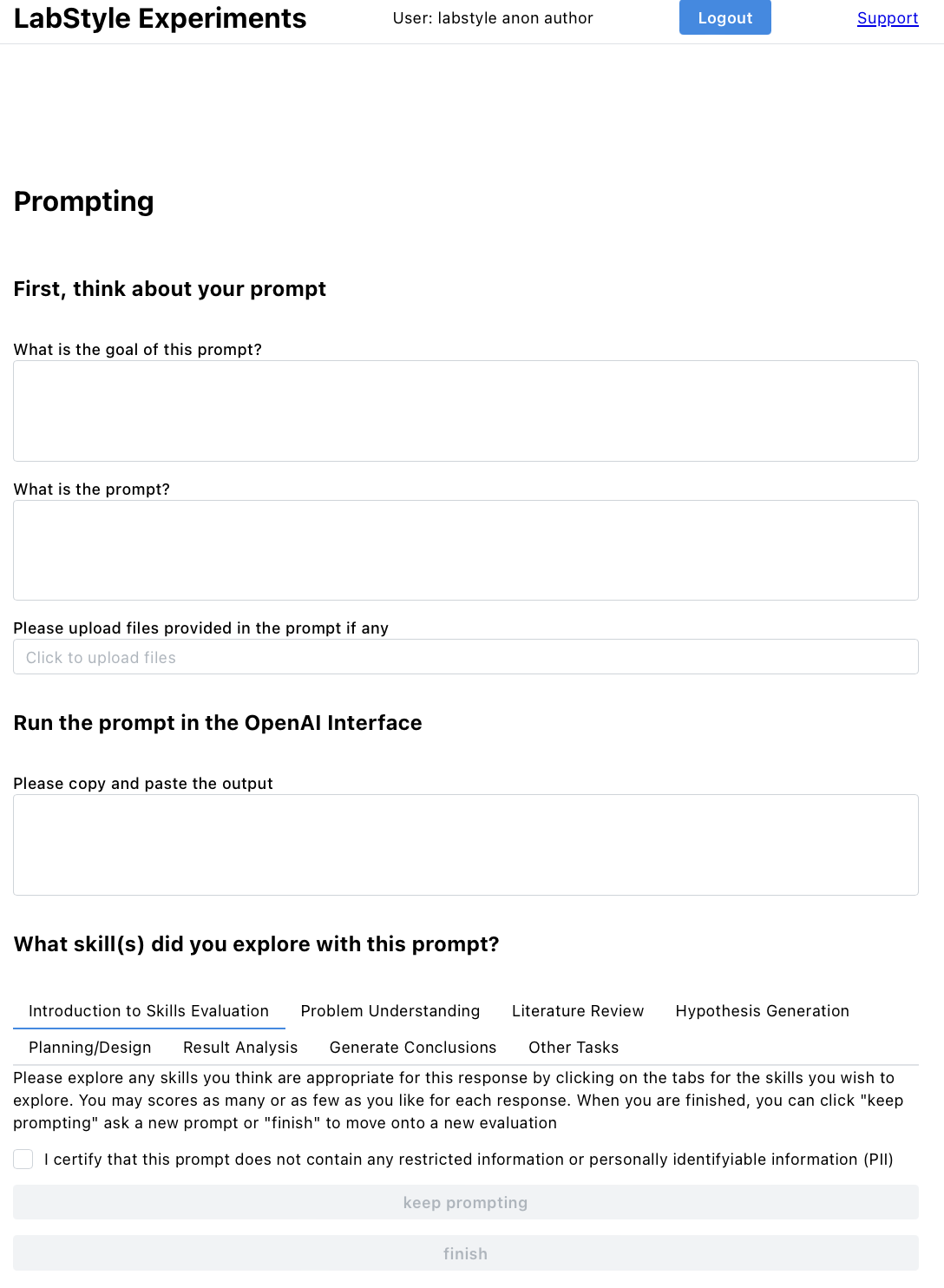}
    \caption{abstyle experiment collection interface: Prompt, response, assessment.}
    \label{fig:interaction}
\end{figure*}

\begin{figure*}[h]
    \centering
    \includegraphics[width=\linewidth]{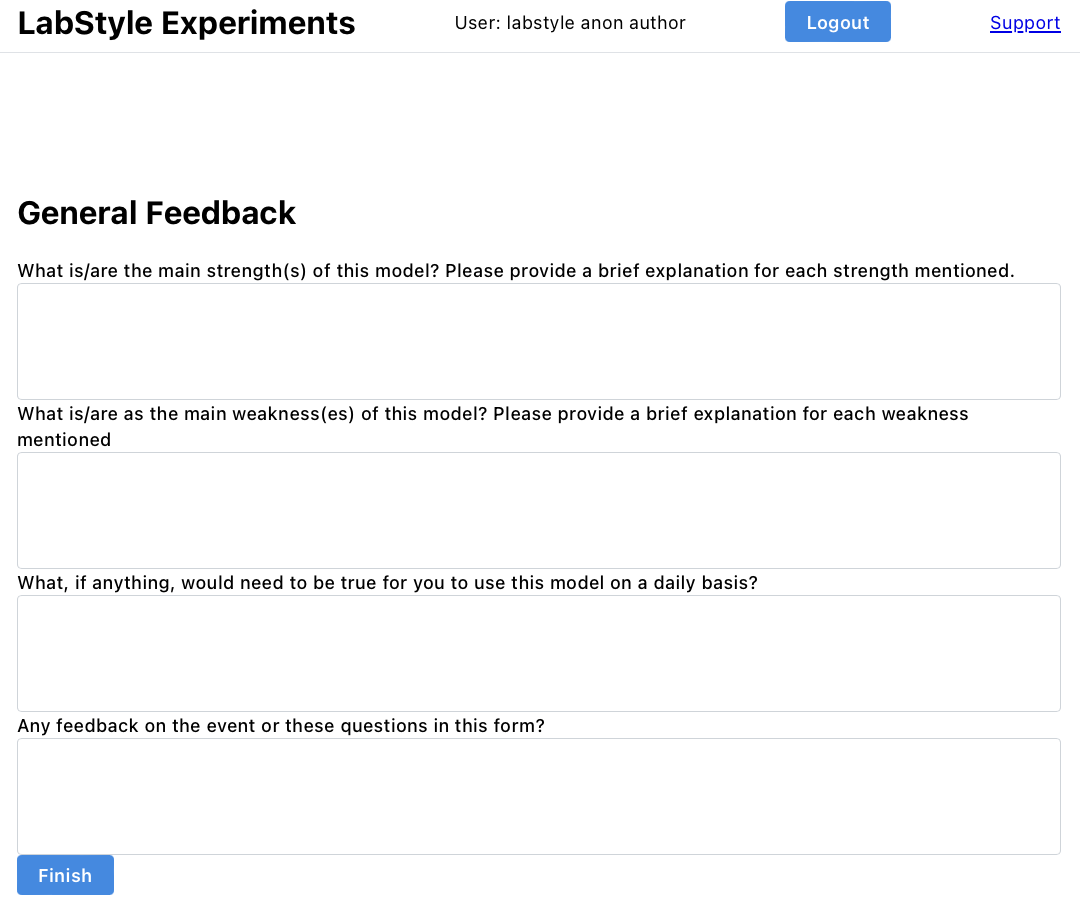}
    \caption{abstyle experiment collection interface: Final assessment.}
    \label{fig:finalassessment}
\end{figure*}

\section{Appendix C: A SciCode benchmark example}
\begin{figure*} 
    \centering
    \includegraphics[width=\textwidth]{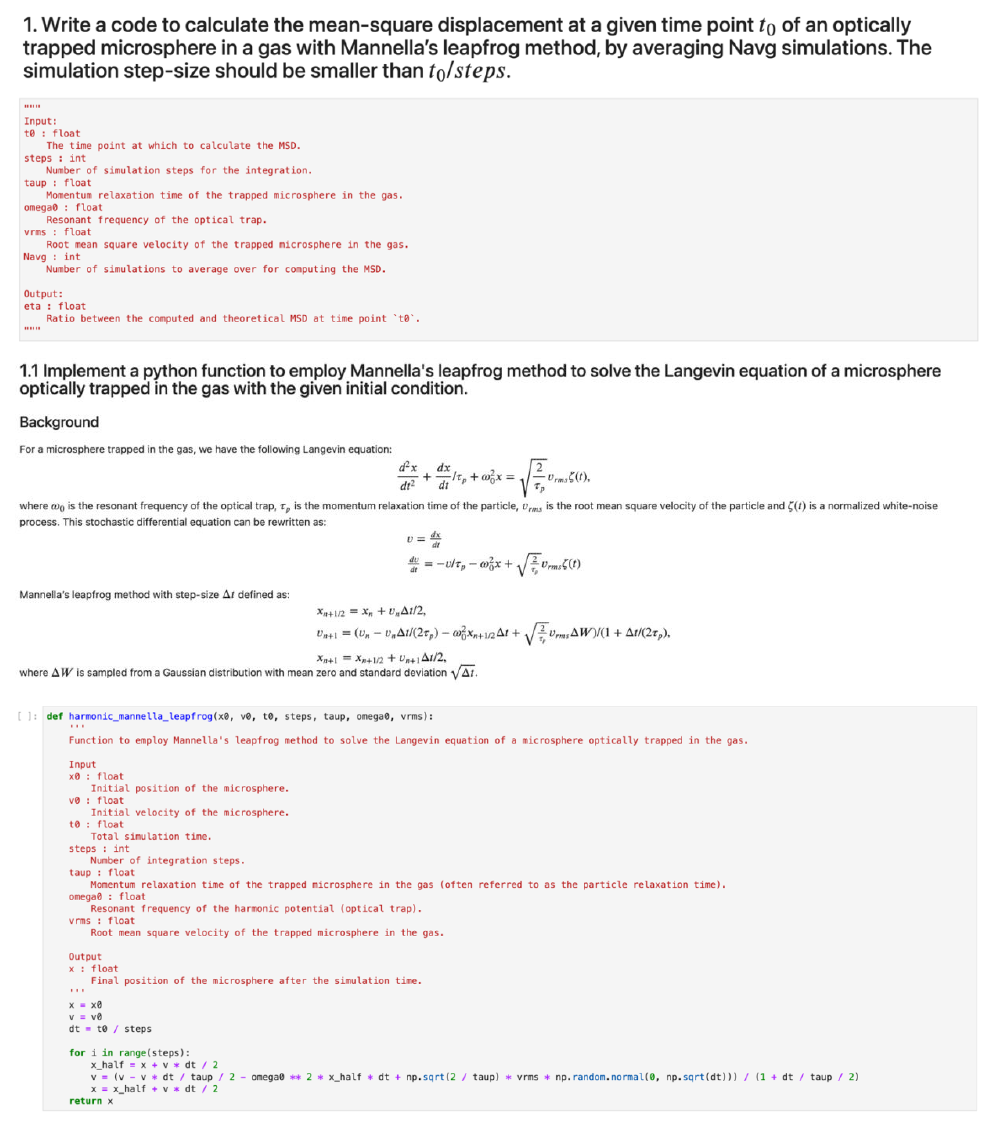} 
    \caption{A SciCode main problem is divided into multiple simpler subproblems for ease of implementation. Docstrings outline the requirements and specify the input-output formats. The scientific background is provided by expert annotators to offer necessary context and guidance.
    }
    \label{fig:scicode_example}
\end{figure*}

\label{app:prompt}

\section{Appendix D: Argo system prompt}
\begin{figure*}[h]
\begin{tcolorbox}[colback=blue!10, colframe=blue!40!black, title=Argo Science Assistant System Prompt]
\small
You are an AI language model named Argo that is a highly knowledgeable AI assistant
specializing in scientific domains such as physics, chemistry, biology, mathematics, and
engineering. Your goal is to provide clear, detailed, and accurate explanations to scientific
questions. Use precise terminology, include relevant equations or formulas when
necessary, and break down complex concepts into understandable parts. Organize your
responses by separating conceptual sections with descriptive sub-header titles to enhance
readability. If applicable, cite credible sources or reference landmark studies to support
your answers.
\end{tcolorbox}
\caption{Argo System Prompt.}
\label{fig:argo-assistant-prompt}
\end{figure*}

\section{Appendix E: ChemRisk system prompt}
\begin{figure*}[h]
\begin{tcolorbox}[colback=blue!10, colframe=blue!40!black, title=Chemical Risk Evaluation System Prompt]
\small
You are an expert in synthetic and computational chemistry with extensive knowledge in organic, inorganic, and organometallic chemistry. Your role is to solve complex chemistry problems by providing accurate and detailed insights into synthesis pathways, reaction mechanisms, chemical properties, and safety considerations. You are well-versed in retrosynthesis, modern synthetic methods, and analytical techniques for structure verification for energetics. Additionally, you excel at interpreting chemical databases and computational predictions to propose efficient and feasible synthetic routes. When responding, ensure that your answers: 1) are concise and actionable, 2) comply with any specified constraints (e.g., have specific atoms or substructures), and 3) produce a chemically valid output in SMILES or SELFIES format.
\end{tcolorbox}
\caption{ChemRisk System Prompt.}
\label{fig:llnl-chem-risk-prompt}
\end{figure*}

\section{Appendix F: LLM as a judge prompt for MCQ evaluation.}
\begin{figure*}[htp]
\clearpage
\begin{tcolorbox}[colback=blue!10, colframe=blue!40!black, title=AI4S Benchmark MCQ LLM-as-a-judge Prompt]
\small
Below is a multiple-choice question, 1 correct answer, 4 incorrect distractors, the domain or field of study, and required skills to answer the question. Be very discriminating, only provide high scores where they are earned, it is crucial to be critical of errors or inadequacies to improve.
Here is the json dictionary formatted multiple choice question, skills and domains:
\begin{verbatim}
{
'Question': '{}',
'Answer': '{}',
'Distractors': {},
'Skills': {},
'Domains': {}
}
\end{verbatim}
Your job is to evaluate the complete question, answers, skills and domain on the following criteria:
\begin{enumerate}
\item Appropriate: Assess whether the question's difficulty aligns with graduate-level knowledge and skills in the subject area. Consider complexity of concepts involved, depth of analysis required, sophistication of language used, application of advanced theories or methodologies. Simple recall from a paper is not sufficiently difficult. Rate the question's appropriateness on a scale of 1--5, where 1 is too basic and 5 is suitably challenging for graduate-level students.
\item Relevant: Evaluate how closely the answer choices relate to the question posed. Consider direct connection between question and answers, absence of extraneous or off-topic information, alignment with the core concept being tested. Score relevance on a scale of 1--5, where 1 indicates poor relevance and 5 indicates high relevance across all answer choices.

\item Complete: Assess whether the answer choices fully address all aspects of the question. Consider coverage of all key elements mentioned in the question, absence of partial or incomplete responses, sufficient detail in each answer choice. There should be one correct answer and four distractors. Rate completeness on a scale of 1--5, where 1 indicates incomplete responses and 5 indicates comprehensive coverage in all answer choices.

\item Correct: Verify that there is only one unambiguously correct answer among the choices. Consider clarity and precision of language in both question and answers, absence of partially correct answers, distinctness of the correct answer from distractors. Score this criterion as either Pass (5) (one clear correct answer) or Fail (0) (multiple correct answers or no correct answer).

\item Controversial: Determine if the correct answer is generally accepted in the field, avoiding contentious or debatable topics. Consider alignment with current academic consensus, avoidance of ongoing debates or unresolved issues, use of well-established facts or theories. Rate the non-controversial nature on a scale of 1-5, where 1 indicates highly controversial and 5 indicates widely accepted, uncontroversial content.

\item Mathematic: Check that the question and answers do not rely on arithmetic calculations. Consider absence of numerical computations, focus on conceptual understanding rather than mathematical operations, use of qualitative rather than quantitative reasoning. Score this criterion as either Pass (no arithmetic required) (5) or Fail (arithmetic is necessary to answer) (0).

\item Skills: Evaluate whether the skills required to answer the question are appropriate for the subject and level. Consider alignment with course learning objectives, relevance to real-world applications in the field, balance of lower-order (recall) and higher-order (analysis, synthesis) thinking skills. Rate the appropriateness of skills on a scale of 1--5, where 1 indicates misaligned skills and 5 indicates perfectly aligned skills for the subject and level.

\item Domains: Assess if the knowledge domains covered by the question are suitable for the subject area. Consider relevance to the course or exam topic, coverage of key subject areas within the field, appropriate breadth and depth of domain knowledge tested. Score the appropriateness of domains on a scale of 1--5, where 1 indicates poorly chosen domains and 5 indicates highly appropriate domains for the subject area.
\end{enumerate}
It is important to be extremely discriminating. Only the best possible questions should receive a maximum score. Correct feedback is vital and preferred over erroneous positivity.
Provide the scores in a json dictionary formatted object with the following fields:
\begin{verbatim}
{
'Appropriate': (score, 'reason'),
'Relevant': (score, 'reason'),
'Complete': (score, 'reason'),
'Correct': (score, 'reason'),
'Controversial': (score, 'reason'),
'Mathematic': (score, 'reason'),
'Skills': (score, 'reason'),
'Domains': (score, 'reason')
}
\end{verbatim}
\end{tcolorbox}
\caption{LLM as a judge prompt for MCQ evaluation.}
\label{fig:ai4s-judge-prompt}
\end{figure*}

\section{Appendix G: LLM as a judge prompt for field style experiment evaluation}
\begin{figure*}[p]
\clearpage
\begin{tcolorbox}[colback=blue!10, colframe=blue!40!black, title=LLM Scientific Reasoning Evaluation Prompt]

\small
You are tasked with analyzing conversation transcripts between humans and a Large Language Model (LLM) to evaluate the LLM's scientific reasoning capabilities. Your objective is to identify the LLM's strengths and weaknesses in various aspects of scientific thinking, using the following framework as a guide. Provide specific examples from the transcript to support your assessment. If a criteria is not applicable to the problem or question being asked in the transcript, note that it is not applicable. Be critical, do not be overly positive if it is not evidenced.

\textbf{Scientific Reasoning Skills Framework}
\textbf{Core Scientific Principles}
\textit{Understanding of the Scientific Method}
\begin{itemize}
\item \textbf{Observation and Questioning:} Does the LLM demonstrate an understanding of how scientific inquiry begins with observation and the formulation of testable questions? Can it identify good vs.\ poorly formed scientific questions?...
\end{itemize}
\textit{Knowledge of Scientific Concepts}
\begin{itemize}
\item \textbf{Domain Knowledge:} Does the LLM possess accurate knowledge of basic scientific concepts in various fields (e.g., biology, chemistry, physics)? How well is it able to answer questions related to different fields of science?...
\end{itemize}
\textit{Critical Evaluation of Scientific Information}
\begin{itemize}
\item \textbf{Source Credibility:} Does the LLM demonstrate an ability to assess the credibility of scientific sources?...
\end{itemize}
\textbf{Specific Scientific Reasoning Skills}
\textit{Experimental Design}
\begin{itemize}
\item \textbf{Identifying Variables:} Can the LLM identify the independent, dependent, and control variables in a given experimental scenario?...
\end{itemize}
\textit{Data Analysis and Interpretation}
\begin{itemize}
\item \textbf{Statistical Significance:} Does the LLM understand the concept of statistical significance?...
\end{itemize}
\textit{Causal Reasoning}
\begin{itemize}
\item \textbf{Identifying Cause and Effect:} Can the LLM correctly identify cause-and-effect relationships in scientific contexts?...
\end{itemize}
\textbf{Communication of Scientific Ideas}
\begin{itemize}
\item \textbf{Clarity and Precision:} Does the LLM communicate scientific ideas clearly and precisely?...
\end{itemize}

\textbf{Scoring Format}

The quantitative assessment should be provided in the following JSON format:

\begin{verbatim}
{
"Core Scientific Principles": {
"Understanding of the Scientific Method": {
"Observation and Questioning": score,
"Hypothesis Formation": score,
"Prediction": score,
"Experimentation": score,
"Data Collection and Analysis": score,
"Conclusion and Theory Formation": score
},
...
}
}
\end{verbatim}

\textbf{Instructions}
\begin{enumerate}
    \item Read the conversation transcript carefully.
    \item Identify instances where the LLM demonstrates strengths or weaknesses in any of the scientific reasoning skills listed above.
    \item For each identified instance, provide:
    \begin{itemize}
        \item The specific skill being assessed (e.g., Hypothesis Formation, Data Analysis: Correlation vs.\ Causation)
        \item A brief description of the context in the conversation
        \item Direct quotes from the transcript that exemplify the LLM's performance (both the user's prompt and the LLM's response)
        \item An assessment of whether this represents a strength or weakness, and a brief explanation of your reasoning
    \end{itemize}
    \item Assign quantitative scores from 1-10 for the criteria as formatted above, if a criteria is not applicable to the transcript give a score of -1.
    \begin{itemize}
        \item A score of -1 means the criteria cannot be assessed as it is not applicable to the transcript
        \item A score of 1 means the LLM completely failed at on the criteria
        \item A score of 10 means the LLM could not have possibly responded better, and completely meets the criteria
    \end{itemize}
\end{enumerate}

\textbf{Transcript}
[Insert transcript here]

\end{tcolorbox}
\caption{LLM as a judge prompt for Field style evaluation.}
\label{fig:jam-judge-prompt}
\end{figure*}
\textbf{Acknowledgment}
The following Argonne researchers voluntarily contributed to the creation of the AI4S benchmarks and the different Lab-style experiments and Field-style experiments.
Adelina Grindeanu
Aikaterini Vriza,
Akshay Dave,
Alec Sandy,
Aleksandr Obabko,
Alex Lavens,
Alex Rodriguez,
Alfonso Napoles,
Allison Bennett-Iron,
Alrhman Abed,
Amaka Okafor,
Andreas Wilke,
Andrei Patapenka,
Andrew Siegel,
Andrey Yakovenko,
Annabelle Boots,
Arvind Ramanathan,
Ayesha Shafiuddin,
Ayman Moawad, 
Azton Wells,
Barnali Chowdry,
Becca Weinberg,
Ben Blakely,
Bipul Barua,
Brahim Mustapha,
Brandon Sforzo,
Brian Ingram,
CD Phatak,
Chandrachur Bhattacharya,
Changyong Park,
Cheng Wang,
Chengjun Sun,
Chiara Bissolotti,
Chihpin Chuang,
Christopher Henry,
Clinton Cohagan,
Dan Meyer,
David Neto,
Debora Meira,
Devesh Reddy,
Dion Antonopoulos,
Doga Gursoy,
Donald Walko,
Eliu Huerta,
Emily Dietrich,
Emily Ohland,
Eshan Sharma,
Fang Zhang,
Fangfang Xia,
Fanny Rodolakis,
Felipe Wang Liu,
Feng Qiu,
Filippo Simini,
Francesco Salucci,
Frank Alexander,
Gautham Dharuman,
Gosia Korbas,
Greg Morin,
Gyorgy Babnigg,
Hairong Shang,
Hanu Arava,
Haoran Wu,
Hassam Harb,
Huihuo Zheng,
Ian Cloet,
Jakob Elias,
James O'Sullivan,
JD Emberson,
Jeffrey Wang,
Jesse Smith,
Jim Grudzinski,
Jiwen Fan,
John Carwardine,
John David Carter,
John Hutchinson,
Jonghwan Kwon,
Jorge Pulpeiro Gonzalez,
Josh Hlavenka,
Juanjuan Huang,
Julie Parente,
Junjing Deng,
Justin Hoffman,
Justin Wozniak,
Kamlesh Suthar,
Katherine Asztalos,
Kathy Macal,
Keith Taddei,
Kent Bostick,
Kent Wootton,
Khalid Hossain,
Kirill Prozument,
Kirsten Laurin-Kovitz,
Krishna Teja Chitty-Venkata,
Kwang Hoon Baek,
Lahsen Assoufid,
Laurent Chapon,
Lee Zachos,
Lisa Childers,
Longwen Ou,
Lorenzo Nocivelli,
Mark Hereld,
Matthew Dearing,
Matthew Diamond,
Matthew Sampson,
Maulik Shukla,
Max Delferro,
Meaghan Bruening,
Megan Clifford,
Mei Zhi-Gang,
Meltem Demirtas,
Meltem Urgun Demirtas,
Michael Buehlman,
Michael Carpenter,
Michael Prince,
Michel van Veenendaal,
Mike Edelen,
Mike Wilkins,
Millie Firestone,
Ming Du,
Minhui Zhu,
Monica Neukomm,
Muhsin Ameen,
Murali Emani,
Murat Keceli,
Mustafa Unal,
Natalia Zuniga,
Nathan Nichols,
Nazib Choudhury Siddique,
Neeraj Hanumante,
Neil James Getty,
Nesar Ramachandra,
Nicholas Frontiere,
Nicholas Lee-Ping Chia,
Nick Goberville,
Nick Moore,
Nicola Ferrier,
Nidhi Gupta,
Nina Andrejevic,
Nithin Manne,
Noah Paulson,
Olaf Borkiewicz,
Olga Antipova,
Osama Mohsen,
Pamela Weisenhorn,
Parfait Gasana,
Peter Kenesei,
Philip Dinemis,
Philippe Piot,
Pinaki Pal,
Prakash Thimmapuram,
Priyash Misra,
Qiaomu Yang,
Quenten Proussard,
Rae Sharp-Geiger,
Rajeev Surendran Assary, 
Rajeev Thakur,
Rajkumar Kettimuthu,
Rao Kotamarthi,
Ravi Madduri,
Rick Vilium,
Rob Ross,
Robert Gaynor,
Roger Sersted,
Rui Hu,
Ryan Aydelott,
Ryan Chu,
Sam Wheeler,
Sang-il Yim,
Scott Collis,
Scott Parent,
Sejal Rhodes
Seth Ockerman
Shelly Kelly,
Shilpika,
Shivam Barwey,
Shubham Kesharwani,
Siba Dowell,
Sibendu Som,
Siby Platthottam,
Sid Raskar,
Simon  Corrodi,
Sixbert Muhoza,
Srutarshi Banerjee,
Stefano Fenu,
Stu Hannay,
Susan Babinec,
Suyin Grass Wang,
Tadbhagya Kumar,
Taemin Kim,
Tanjin He,
Tanwi Mallick,
Taylor Childers,
Tekin Bicer,
Temitope Oproudek,
Tianhao Gu,
Tianyi Li,
Tiffany Kinnibrugh,
Tim Ashenfelter,
Tim Hobbs,
Tim Nguyen,
Tim Williams,
Timothy Suzuki,
Todd Munson,
Tom Brettin,
Tom Peterka,
Tom Uram,
Troy Arcomano,
Utkarsh Nanda
Valerie Taylor,
Varuni Sastry,
Venkata Devesh Reddy, Seethi
Vincent Chirio,
Vincent Chia
Vincenzo Cappello,
Vivian Sullivan,
Vrindaa Somjit,
Wei Li,
Weixing Cheng,
Wen Zhuang,
Xiaoyang Liu,
Xuli Wu,
Yang Wang,
Yanna Chen,
Yawei Yang,
Yeni  Li,
Yeonjun Jeong,
Yildiz Ocrun,
Yinghu Piao,
Yiqi Yu,
Yue Cao,
Yuejun Yan,
Yuepeng Zhang,
Yuri Oksuzian,
Zhao Zixuan,
Zhi Zhou,
Zixuan Zhao.
\end{document}